%% file: sn-article.tex
\documentclass[pdflatex,sn-apa]{sn-jnl}

\usepackage{graphicx}%
\usepackage{multirow}%
\usepackage{amsmath,amssymb,amsfonts}%
\usepackage{amsthm}%
\usepackage{mathrsfs}%
\usepackage[title]{appendix}%
\usepackage{xcolor}%
\usepackage{textcomp}%
\usepackage{manyfoot}%
\usepackage{booktabs}%
\usepackage{algorithm}%
\usepackage{algorithmicx}%
\usepackage{algpseudocode}%
\usepackage{listings}%
\usepackage{graphicx}
\usepackage{subcaption}
\usepackage{array}
\usepackage[T1]{fontenc}
\usepackage[utf8]{inputenc}
\usepackage{tabularx}

\newif\ifanonymous
\anonymoustrue   

\theoremstyle{thmstyleone}%
\theoremstyle{thmstyletwo}%

\theoremstyle{thmstylethree}%

\begin{document}

\title[GenAI VP Logs to Process Evidence]{From GenAI Virtual Patient Dialogue Logs to Teacher-Interpretable Process Evidence: A Learning Analytics Study in Higher Education}








\author[1]{\fnm{Xinyu} \sur{Li}}\email{xinyu.li1@monash.edu}
\author[2]{\fnm{Zijian} \sur{Li}}\email{zijianli@stu.pku.edu.cn}
\author[2]{\fnm{Mengyu} \sur{Xia}}\email{marine\_xia@stu.pku.edu.cn}
\author[2]{\fnm{Luzhen} \sur{Tang}}\email{tangluzhen@pku.edu.cn}
\author[3]{\fnm{Naping} \sur{Chen}}\email{s\_chennp@stu.edu.cn}
\author[4]{\fnm{Changmin} \sur{Lin}}\email{cocolin@stu.edu.cn}
\author[6,7]{\fnm{Danijela} \sur{Gašević}}\email{danijela.gasevic@monash.edu}
\author*[5,1]{\fnm{Dragan} \sur{Gašević}}\email{dgasevic@hku.hk}
\author*[2]{\fnm{Yizhou} \sur{Fan}}\email{fyz@pku.edu.cn}

\affil[1]{\orgdiv{Faculty of Information Technology}, \orgname{Monash University}, \orgaddress{ \city{Melbourne}, \postcode{3800}, \country{Australia}}}

\affil[2]{\orgdiv{Graduate School of Education}, \orgname{Peking University}, \orgaddress{\city{Beijing}, \postcode{100871}, \country{China}}}

\affil[3]{\orgdiv{Department of Clinical Skills Training Center,}, \orgname{Shantou University Medical College}, \orgaddress{\city{Shantou}, \postcode{515041}, \country{China}}}

\affil[4]{\orgdiv{Office of Teaching Affairs}, \orgname{Shantou University Medical College}, \orgaddress{\city{Shantou}, \postcode{515041}, \country{China}}}

\affil[5]{\orgdiv{Faculty of Education \& School of Computing and Data Science}, \orgname{The University of Hong Kong}, \orgaddress{\city{Hong Kong}, \country{China}}}

\affil[6]{\orgdiv{School of Public Health}, \orgname{The University of Hong Kong}, \orgaddress{\city{Hong Kong}, \country{China}}}

\affil[7]{\orgdiv{School of Public Health and Preventive Medicine}, \orgname{Monash University}, \orgaddress{\city{Melbourne}, \postcode{3004}, \country{Australia}}}


\abstract{Medical history taking is a dialogue-based clinical reasoning task in which learners must gather, organise, and integrate patient information while the consultation unfolds. Generative AI-powered virtual patients (GenAI VPs) make repeated history taking practice scalable and preserve full turn by turn dialogue. However, these logs are educationally difficult to use directly. Complete transcripts are too detailed for routine teacher review, whereas final scores obscure whether learners followed up patient cues, checked uncertainty, or used summaries to guide later questioning. This study examined whether coded GenAI VP dialogues can provide teacher-interpretable process evidence of clinical reasoning. We analysed 1{,}030 GenAI VP dialogues from 210 second-year medical learners across five weeks chest-pain cases. Each consultation was teacher-scored using a rubric assessing the full history taking dialogue, and consultations were classified within each week as high- or low-rated using the weekly median score. To explain how rated performance was reflected in the dialogue process, we applied three analytic layers to the same coded dialogue data: behavioural prevalence, local co-occurrence using Epistemic Network Analysis, and sequential transition using Transition Network Analysis. High-rated consultations involved more history taking activity, but differences were not simply about volume. High rated consultations more often connected information gathering and symptom exploration with communication, checking, organisation, and synthesis. Summarising and organising moves more often led to verification or mechanism-oriented follow-up. These findings show how layered analysis of GenAI VP dialogue logs can reveal process patterns associated with high rated history taking and support process-focused feedback in medical education.}




\maketitle

\input{sections/01-introduction}

\input{sections/02-literature-review}
\input{sections/03-methods}
\input{sections/04-results}

\input{sections/05-discussion}

\input{sections/06-conclusion}

\backmatter





\subsection*{Abbreviations}

\section*{Declarations}
Identifying details (author names, affiliations, funding sources, ethics approval numbers, and author contributions) are anonymised in the submitted manuscript for double-blind peer review and will be provided on the title page and in the final manuscript.

\subsection*{Acknowledgements}
The authors thank the participating students and instructors for their contributions to the study.

\subsection*{Funding}
Blinded for review.

\subsection*{Conflict of interest}
The authors declare that they have no competing interests.

\subsection*{Ethics approval and consent to participate}
The study was approved by the institutional ethics committee at the participating medical college (specific approval details anonymised for double-blind peer review and provided on the title page). All participants received written information about study objectives, procedures, and participant rights, and provided written informed consent prior to participation.

\subsection*{Data Availability}
The datasets used and analysed during the current study are available from the corresponding author on reasonable request.

\subsection*{Authors Contribution}
Blinded for review.




\bigskip
\begin{flushleft}%
Editorial Policies for:

\bigskip\noindent
Springer journals and proceedings: \url{https://www.springer.com/gp/editorial-policies}

\bigskip\noindent
Nature Portfolio journals: \url{https://www.nature.com/nature-research/editorial-policies}

\bigskip\noindent
\textit{Scientific Reports}: \url{https://www.nature.com/srep/journal-policies/editorial-policies}

\bigskip\noindent
BMC journals: \url{https://www.biomedcentral.com/getpublished/editorial-policies}
\end{flushleft}

\ifanonymous
  \bibliography{bibliography_anonymised}
\else
  \bibliography{sn-bibliography}
\fi

\end{document}

%% file: sections/01-introduction.tex
\section{Introduction}\label{sec1}

History taking is a central component of medical education because it requires learners to conduct a clinical conversation with a patient while simultaneously reasoning about the patient's problem. During a consultation, learners must gather relevant information, recognise cues, clarify uncertainty, adjust questioning, and build a coherent account of the patient's condition \citep{shea2023clinical,windish2005teaching}. Clinical reasoning therefore does not occur only when a diagnosis is stated at the end of a case; it unfolds throughout the consultation as learners judge which information matters, which symptoms require follow-up, and whether the information gathered so far supports a candidate explanation \citep{eva2005every}. In this sense, history taking is a form of clinical reasoning carried out through dialogue.

Assessing clinical reasoning as a process remains difficult. Course assessment and clinical skills assessment often rely on final grades, task outcomes, checklist scores, diagnostic accuracy, or overall ratings \citep{black1998assessment,bennett2011formative}. These measures can indicate whether a learner met a standard, but they reveal less about how the learner reached that outcome. In a chest pain case, for example, two learners may arrive at a similar final diagnosis, but differ substantially in whether they followed up pain characteristics, checked risk factors, clarified ambiguous answers, connected symptoms across domains, or summarised information at appropriate moments \citep{bickley2012bates,hasnain2001historytaking}. These in-process behaviours are important because they make metacognitive regulation partly visible: the learner must monitor what is known, notice what remains uncertain, and decide how the next question should move the consultation forward \citep{goldowsky2023self}.

Providing repeated opportunities to practise and assess such processes is challenging in medical education. Standardised patient (SP) encounters and objective structured clinical examinations (OSCEs) offer authentic practice and structured assessment, but they require trained personnel, scheduling, room arrangements, and safeguards for rating consistency \citep{hamilton2024evolution}. These logistical demands limit how much complete-consultation practice each learner can receive, especially in large cohorts \citep{plackett2022effectiveness}. Even when SP or OSCE encounters are recorded, the recordings are rarely transformed into turn-by-turn evidence that educators can use for routine feedback on learners' reasoning processes \citep{wagner2020communication}.

Generative AI-powered virtual patients (GenAI VPs) offer a way to address limitations in existing practice and evidence generation in history taking education. Unlike scripted or menu-based virtual patients, a GenAI VP allows learners to ask questions freely in natural language and generates patient responses within the frame of a clinical case \citep{yi2025feasibility}. As a result, the same case can unfold into different consultations depending on each learner's questioning path \citep{holderried2024generative}. This makes repeated, large-scale history taking practice more feasible and, importantly, preserves the full learner--GenAI VP dialogue as turn-by-turn data for later analysis \citep{li2026large}.

This new source of dialogue data also creates an assessment problem. The educational challenge is therefore not simply that GenAI VP systems generate large amounts of dialogue data. The challenge is that the two most accessible forms of evidence are both limited for formative use. A final score can indicate whether a consultation was judged successful, but it cannot show where the learner missed a cue, repeated already answered questions, failed to clarify uncertainty, or used a summary to redirect later questioning. A full transcript contains this information, but it is too detailed for routine inspection by teachers in large cohorts and repeated practice tasks \citep{swiecki2022assessment,molenaar2023measuring}. For GenAI VP practice to support clinical reasoning education, dialogue logs therefore need to be transformed into process evidence that is detailed enough to show how history taking unfolded, but structured enough for teachers to interpret.

This need motivates the present study. Rather than treating GenAI VP logs as raw transcripts or reducing them to final scores, we examined whether coded learner turns could reveal performance-related patterns in history taking processes. The comparison between high- and low-rated consultations was used as a performance-anchored contrast \citep{thampy2019assessing,haring2020validity}. It was not intended to classify learners as generally strong or weak. Instead, teacher-rated history taking scores were used to examine whether consultations judged to be of high or low quality within the same weekly case showed different dialogue processes. This within-week comparison was important because the five cases differed in clinical content and likely difficulty; comparing learners within the same weekly task allowed process differences to be interpreted against the same case and scoring context.

To make these process differences interpretable, we applied a layered analytic approach to the same coded dialogue data. Behavioural prevalence analysis examined which coded history taking behaviours occurred and how often they appeared. Local co-occurrence analysis examined which behaviours were connected within nearby learner turns. Sequential transition analysis examined which behaviours tended to follow one another as the consultation unfolded. These three layers address complementary educational questions: whether important behaviours were present, whether they were locally coordinated with other clinically relevant moves, and whether they guided the next step of the consultation \citep{matcha2019analytics,shaffer2016tutorial,saqr2024sequence}. The present study analysed GenAI VP history taking tasks completed by 210 second-year medical learners over five weeks. The analysis drew on a coding scheme for GenAI VP questioning behaviours developed in earlier work \citep{chen2026developing}.

%% file: sections/02-literature-review.tex
\section{Literature Review}\label{sec2}

\subsection{History taking as observable metacognitive regulation of clinical reasoning}\label{sec:hist_regulation}

History taking is an early and observable setting in which medical learners practise clinical reasoning \citep{keifenheim2015teaching}. During history taking, learners gather information through patient dialogue, elicit symptom details, clarify ambiguity, follow clinically relevant cues, and organise information for later diagnostic and management decisions \citep{mahbubani2023history}. History taking is therefore not a checklist of symptom questions. It is a dialogue-based inquiry practice in which learners decide what to ask, how to respond to patient answers, and when to reorganise collected information while the consultation is still unfolding \citep{xu2021methods}.

Clinical reasoning theories explain why the organisation of history taking matters \citep{bowen2006educational,charlin2007scripts}. Hypothetico-deductive accounts describe how early patient information triggers provisional explanations that later questions test and refine \citep{si2022strategies}. Illness-script and knowledge-encapsulation accounts emphasise organised case-based knowledge that helps clinicians connect symptoms, risk factors, mechanisms, and likely diagnoses \citep{schmidt2007expertise}. Across these accounts, competent clinical reasoning depends on metacognition: monitoring what is known, noticing what remains incomplete or uncertain, and adjusting inquiry accordingly \citep{cutrer2017fostering}. In history taking, these metacognitive demands can be reflected in observable behaviours such as systematic coverage, cue-responsive follow-up, clarification, hypothesis-sensitive questioning, checking, and interim summarising \citep{ng2025clinical}.

The metacognitive character of history taking makes records of dialogue in patient consultations useful for process analysis \citep{winne2022modeling}. Dialogue records cannot directly reveal metacognitive states such as planning, monitoring, or uncertainty regulation \citep{rakovic2023harnessing}. Dialogue records can, however, preserve visible counterparts of these functions: how learners organise questions, respond to cues, confirm ambiguity, integrate information, and move between routine questioning and reasoning-oriented moves \citep{winne2022modeling}. History taking dialogue therefore provides observable behavioural traces of clinical reasoning, even though these traces are not direct measurements of cognition \citep{veenman2007assessment}.

\subsection{From simulated encounters to analysable GenAI VP dialogue corpora}\label{sec:genai_changes}

Simulation-based formats have long supported history taking practice and assessment in medical education. SP encounters and OSCEs provide interactive clinical scenarios and structured opportunities to judge learner performance \citep{malauaduli2022osce}. For process analysis of history taking, however, the key issue is not whether these formats are educationally valuable, but whether they can routinely generate fine-grained, comparable, and reusable process data at scale. In many teaching settings, the interaction itself remains difficult to convert into turn-by-turn evidence for everyday feedback and research, even when performance ratings or recordings are available \citep{misra2024osce,wagner2020communication}.

Earlier virtual patient systems addressed some problems of standardisation and accessibility by allowing learners to work through repeatable clinical cases \citep{faferek2024integrating}. These systems also made certain learner actions easier to record, such as menu selections, chosen pathways, time on task, and diagnostic decisions \citep{kononowicz2019virtual}. Yet many earlier VP designs constrained learner history taking through predefined options or scripted branches \citep{jay2025use}. For instance, branched VP systems such as those used in the TAME project structured interactions around an "ideal" pathway of patient management decisions, with linear variants offering no deviation from scripted routes \citep{woodham2019virtual}. Similarly, VPs commonly assessed clinical reasoning through multiple-choice questions or discrete decision points, while the non-linearity of clinical reasoning posed a persistent challenge for scoring and feedback that quantitative methods could not sufficiently capture \citep{hege2018advancing}. Such designs supported consistency, but they offered limited evidence about how learners formulate their own questions, redirect inquiry after patient responses, or reorganise information during an open-ended consultation \citep{maicher2023artificial}.

GenAI VPs introduce a different data affordance for history taking research and education. Because learners can ask questions in natural language and receive case-bounded patient-like responses, the same clinical case can produce many learner-generated consultation paths \citep{yi2025feasibility,brugge2024large}. This changes the data analysis focus from a fixed pathway or final decision to a corpus of complete learner--VP dialogues. These logs make it possible to compare how learners working on the same clinical problem initiate inquiry, pursue cues, check uncertainty, and reorganise information over time \citep{holderried2024generative,holderried2024feedback}. In this sense, GenAI VPs change not only the conditions of practice, but also the conditions under which history taking processes can be studied \citep{bond2024meta,li2026large}.

\subsection{Learning analytics from dialogue records to process evidence}\label{sec:la_process_evidence}

Dialogue records do not automatically become usable evidence for assessment or feedback \citep{swiecki2022assessment}. A coded learner turn does not have a fixed educational meaning in isolation. Its meaning depends partly on the utterances immediately before and after it within the same consultation \citep{reimann2007time}. For example, a symptom-specific question may introduce a new line of clinically relevant line of questioning, clarify an ambiguous patient answer, or repeat information that the patient has already provided \citep{nendaz2006beyond}. The same coded behaviour can therefore carry different meanings depending on the behaviour's position in the dialogue \citep{sonnenberg2015discovering}. Because the evidence is sequential and interactional, coding individual utterances is only a first step. Additional analytic methods are needed to examine how coded behaviours accumulate, combine, and unfold over time. In this sense, learning analytics provides a methodological bridge between raw dialogue records and interpretable process evidence for teachers \citep{verbert2014learning}. Learning analytics is therefore needed to transform dialogue records into interpretable process evidence rather than treating raw transcripts as self-explanatory \citep{li2023analytics}.

Teacher-rated history taking rubric scores provide a necessary performance anchor because they summarise the assessed quality of the full consultation dialogue. However, such scores do not explain the dialogue process that produced the rating \citep{thampy2019assessing}. A history taking score may reflect broader coverage, better organisation, more appropriate follow-up to patient cues, or more effective use of summaries and checks \citep{haring2020validity}. Similar scores may also conceal different history taking behaviour paths \citep{regehr1998comparing}. For example, two learners may receive similar history taking scores because both covered the required domains. Yet one learner may first characterise the chest pain, follow up a patient cue about exertion, summarise the emerging pattern, and then check risk factors, whereas another learner may ask the same domains as a checklist and repeat information already given. The score may be similar, but the dialogue process differs. Linking teacher-rated performance to coded dialogue behaviour is therefore necessary if GenAI VP logs are to support formative feedback rather than only summative classification \citep{matcha2019analytics}. Interaction coding provides one route into this linkage \citep{roter2002roter}. Consultation frameworks such as the Calgary-Cambridge Guide \citep{kurtz2003marrying}, the Roter Interaction Analysis System \citep{roter2002roter}, classify communicative and structural features of medical encounters. In addition, history taking assessment work has identified observable indicators of clinical reasoning, including recognising relevant information, specifying symptoms, asking pathophysiologically oriented questions, putting questions in a logical order, checking with the patient, and summarising \citep{haring2017observable,furstenberg2020assessing}

The unresolved problem is how to connect coded dialogue behaviours with assessed consultation quality without losing the sequential character of the consultation. A count-based view can identify whether behaviours such as symptom specification, checking, logical organisation, or summarising appear more often in consultations that teachers judge to be of higher quality. This is educationally useful because assessment and communication frameworks treat these behaviours as relevant indicators of history taking quality and clinical reasoning during the encounter \citep{kurtz2003marrying,roter2002roter,haring2017observable,furstenberg2020assessing}. However, counts alone do not show how those behaviours are positioned around other learner moves, and prior learning analytics research has shown that process evidence requires attention to how actions are situated in time rather than only how often they occur \citep{reimann2007time,matcha2019analytics,swiecki2022assessment}. A routine question placed near checking, summarising, or logical organisation may contribute to a coherent clinical account, whereas a routine question placed mainly near further routine questions may reflect checklist-like coverage. Local coordination also does not show temporal direction. A summary followed by checking or mechanism-oriented questioning suggests a different consultation process from a summary followed by repeated questioning, even though both cases contain the same summary code. ENA and sequence-oriented learning analytics provide ways to examine these differences because they model, respectively, local connections among coded behaviours and the order in which coded behaviours unfold \citep{shaffer2016tutorial,csanadi2018coding,saint2020combining,saint2022temporally,tao2025exploring}. These limitations create three related gaps for GenAI VP history taking research: limited evidence about which coded behaviours are associated with teacher-rated consultation quality, limited evidence about how those behaviours are locally coordinated within short episodes of a consultation, and limited evidence about how one coded behaviour leads into the next step of the consultation. Comparing high- and low-rated consultations is therefore useful when it is treated as a performance-anchored contrast rather than as a claim about stable learner ability. Such a contrast can show whether consultations judged under the same case and scoring context differ in behavioural prevalence, local co-occurrence, and sequential transition \citep{thampy2019assessing,haring2020validity,li2023analytics}.

\subsection{Research Questions}

Building on these gaps, the study used teacher-rated history taking rubric scores as a performance anchor for analysing GenAI VP dialogue processes. High- and low-rated consultations were compared within each weekly case, not to classify learners as generally strong or weak, but to examine whether consultations judged under the same case and scoring context differed in observable history taking processes. This study asks the three following research questions:

\begin{itemize}
    \item \textbf{RQ1 (Behavioural prevalence):} Which coded history taking behaviours distinguish high- and low-rated GenAI VP history taking consultations?

    \item \textbf{RQ2 (Local co-occurrence):} How are coded history taking behaviours combined in nearby turns in high- and low-rated consultations?

    \item \textbf{RQ3 (Sequential transitions):} How do high- and low-rated consultations differ in transitions from one coded history taking behaviour to another?
\end{itemize}

%% file: sections/03-methods.tex
\section{Methods}\label{sec:methods}

\subsection{Participants}
A total of 210 second-year medical students from an [anonymised] medical college in China participated from April to May 2024. At the time of the study, these students were enrolled in the \textit{Symptomatology and History Taking} course and had completed core instruction on foundational history taking knowledge (e.g., key history domains, basic interviewing approaches, and selected symptom-focused content).

Analyses were conducted separately for each week using all valid dialogue logs available for the weekly task. After integrity checks at the task level (see Data preparation), the analytic samples were: W1 $n=206$, W2 $n=207$, W3 $n=205$, W4 $n=206$, and W5 $n=206$. In total, 197 students had complete dialogue logs for all five weeks (W1 to W5). 

\subsection{Study design and learning environment}
\subsubsection{Ethics}
The study was approved by the institutional ethics committee at the participating medical college (specific approval details anonymised for double-blind peer review). Participants received written information about study objectives, procedures, and participant rights, and provided informed consent prior to participation. Participants were informed that they could withdraw at any time; upon withdrawal, their data would be removed from the study records.

\subsubsection{Learning environment and tasks}
All tasks were completed on [Anonymous system], a Moodle integrated learning platform that hosted instructional resources (medical knowledge, diagnostic principles, disease reasoning guidance, and evaluation criteria) and integrated tools supporting consultation and submission workflows \citep{li2026flora}. The GenAI VP was implemented as a chatbot in [Anonymous system]. The chatbot used a case-specific prompt designed for GPT-3.5 that specified the patient role, the clinical history of the case, and response constraints intended to keep the virtual patient in role during the consultation. The platform recorded the full learner-VP dialogue for subsequent process analysis. The five weekly tasks were chest pain cases representing spontaneous pneumothorax, stable angina, aortic dissection, acute pulmonary embolism, and acute pericarditis. In each task, learners conducted a history taking encounter with the GenAI VP, could take notes during the encounter, and submitted a diagnostic conclusion at the end of the task.

\subsubsection{Task procedure}
The procedure relevant to the present study comprised an orientation and training task followed by five consecutive history taking tasks. Before the history taking tasks, participants reviewed task instructions and rubrics and watched an instructional demonstration video. Each task followed the same structure: participants reviewed instructions and criteria, conducted the history taking with the GenAI VP while optionally taking notes and consulting instructional resources, and then drafted and submitted a diagnostic conclusion. The analyses reported in this paper used the history taking dialogue logs and weekly teacher-rated history taking scores.

\subsubsection{History taking performance scores}
The performance variable used for grouping was the weekly history taking score assigned to each consultation dialogue. The score was based on the full consultation dialogue rather than the submitted diagnosis alone. The scoring rubric was developed using the Kalamazoo Essential Elements checklist for medical encounters \citep{makoul2001essential}, national medical licensing examination criteria for history taking, and course teaching requirements. The rubric covered history information, including chief complaint and history of present illness, as well as communication techniques, such as use of medical terminology and continuity of questioning \citep{milota2019narrative}. It recorded two components: a history taking content score (HT-Content) and a history taking technique score (HT-Technique), which together formed the total history taking score (HT-Total).

To examine scoring consistency, 20 participants were randomly selected and their five history taking exercises were scored independently by three raters: two course instructors and one fifth-year medical student. The raters worked independently and were blinded to one another's ratings. Because the rubric contained multiple ordinal or score-based items, inter-rater consistency was summarised with Cronbach's alpha for the rubric components and total score rather than reporting every item separately. For each weekly task, Cronbach's alpha was calculated across the three raters' scores for HT-Content, HT-Technique, and HT-Total. Across the five weekly tasks, the total history taking score showed high consistency ($\alpha=.926$--$.966$); the content component ranged from $\alpha=.887$ to $.964$, and the technique component ranged from $\alpha=.753$ to $.929$. The final diagnostic conclusion was treated separately as correct or incorrect and was not used to define the high and low performance groups in this paper.

\subsection{Dialogue dataset preparation}

To protect privacy, identifiable information was removed and replaced with de-identified user IDs. Names, contact details, and other direct identifiers were excluded from the analytic dataset. Data integrity checks were performed at the task level to ensure that each included record contained a complete consultation dialogue log required for process analysis. Because analyses were conducted separately for each week, inclusion was determined week by week: a learner contributes to week $t$ if their dialogue log for week $t$ meets completeness criteria. The resulting analytic sample sizes varied slightly across weeks (W1 $n=206$, W2 $n=207$, W3 $n=205$, W4 $n=206$, W5 $n=206$). A subset of 197 learners had complete dialogue logs for all five weeks.

To transform raw dialogue into interpretable indicators of clinical reasoning and history taking processes on the learner side, learner utterances were represented using the behavioural coding scheme developed and validated in prior work on GenAI VP history taking dialogues \citep{chen2026developing}. The prior coding study developed a 12-code scheme across three behavioural dimensions and established inter-rater reliability through iterative calibration, with Cohen's $\kappa$ calculated on independent pre-discussion ratings and all codes reaching the predefined reliability threshold. The coding scheme operationalised task-relevant learner dialogue behaviours, such as questioning driven by hypotheses, follow-up to patient cues, clinically grounded organisation, and integrative synthesis. In the present study, we used the finalised coded dialogue dataset as the basis for process analysis. The coding unit was the learner utterance. Patient responses generated by the GenAI VP were retained as the interactional context for learner questioning but were not treated as behavioural states in the data analyses. Table~\ref{tab:codebook} summarises the codebook used for the present analysis.

\begin{table}[ht]
\centering
\caption{Codebook for identifying learners' behaviours in history taking dialogues.}
\label{tab:codebook}
\renewcommand{\arraystretch}{1.3}  
\begin{tabular}{p{0.15\textwidth}p{0.33\textwidth}p{0.44\textwidth}}
\hline
\textbf{Dimension} & \textbf{Code Name (Abbreviation)} & \textbf{Operational definition (brief)} \\
\hline
\multirow{4}{*}{\parbox{0.15\textwidth}{Clinical\\ Reasoning}}
  & Pathophysiological Question (PQ) & Hypothesis-driven or mechanism-oriented questioning with clear diagnostic intent. \\
  & Relevant Response (RR) & Follow-up that recognises and probes diagnostically significant cues introduced by the patient. \\
  & Summarising \& Integrating (SI) & Synthesis and restructuring of collected information into a coherent summary that supports reasoning. \\
  & Logical Organisation (LO) & Clinically coherent organisation of inquiry based on relevance to the chief complaint and differential value. \\
\hline
\multirow{6}{*}{\parbox{0.15\textwidth}{Information\\ Gathering}}
  & Specifying Symptoms (SS) & Systematic probing of symptom characteristics (e.g., onset, duration, severity, triggers). \\
  & Routine Question (RQ) & Standard history taking questions (e.g., basic history domains, routine openers) resembling interview checklists. \\
  & Summarising \& Restating (SR) & Restating collected information without restructuring or diagnostic synthesis. \\
  & Checking (CK) & Clarifying or confirming patient-reported information (e.g., resolving ambiguity or inconsistency). \\
  & Repeating Question (RT) & Redundant or repeated questioning targeting the same information point. \\
  & Fuzzy Question (FQ)  & Repeated vague or overly broad open-ended prompts within a domain (beyond appropriate initial openers). \\
\hline
\multirow{2}{*}{\parbox{0.15\textwidth}{Communication}}
  & Facilitative Communication (CC) & Greetings, reassurance, brief explanations, transitional phrases, and other non-diagnostic social talk. \\
  & Off-topic Statement (OS) & Illogical, incomplete, or case-irrelevant utterances that cannot be meaningfully classified elsewhere. \\
\hline
\end{tabular}
\end{table}

The codes in Table~\ref{tab:codebook} operationalise the learner-side behaviours through which clinical reasoning is enacted in a history taking dialogue. As established in Section~\ref{sec:hist_regulation}, history taking is not treated in this study as a checklist of questions, but as a dialogue-based clinical reasoning process in which learners gather information, recognise patient cues, clarify ambiguity, organise what has been established, and decide how later questions should develop the consultation \citep{bowen2006educational,charlin2007scripts,cutrer2017fostering}. The coding scheme therefore provided an utterance-level representation of observable history taking behaviours, including information gathering, symptom specification, checking, cue-responsive follow-up, logical organisation, mechanism-oriented questioning, communication management, and summarising \citep{haring2017observable,furstenberg2020assessing}.

The analyses did not recode these utterances as metacognitive states. This distinction is important because dialogue logs record what learners said in the consultation, not what they were consciously planning, monitoring, or regulating. For that reason, all statistical analyses used the original behavioural codes and their observed frequency, co-occurrence, and transition patterns. When interpreting the results, we related selected patterns involving codes such as LO, SI, CK, RR, and PQ to the clinical reasoning and metacognitive functions reviewed in Section~\ref{sec:hist_regulation}; however, these interpretations were treated as behavioural inferences from dialogue traces rather than direct measurements of learners' cognition \citep{winne2022modeling,rakovic2023harnessing}.

\subsection{Analysis plan}


All analyses were conducted week by week (W1 to W5). This decision was aligned with the research questions, which asked whether consultations judged as high or low in quality within the same GenAI VP case showed different behavioural prevalence, local co-occurrence, and sequential transition patterns. The study did not aim to estimate longitudinal growth across the five weeks. Instead, each weekly task was treated as a separate case context because the five chest-pain cases differed in clinical content and likely difficulty. Analysing weeks separately therefore kept each high--low contrast anchored to learners working on the same case and being judged within the same weekly scoring context.

Within each week, learners were split into a high-rated and a low-rated group using that week's teacher-rated total history taking score. Learners at or above the weekly median formed the high-rated group, and learners below the weekly median formed the low-rated group. Because the grouping was performed separately by week, group membership could vary across weeks. The groups should therefore be interpreted as consultation-level performance contrasts within a weekly case, not as stable learner-level ability groups. This grouping strategy allowed the same performance-anchored comparison to be used across the three analytic layers: behavioural prevalence for RQ1, local co-occurrence for RQ2, and sequential transition for RQ3. Prior to analysis, the Off Topic Statement (OS) category was excluded from the dialogue data because it was extraneous to the research objectives.

\subsubsection{Behavioural prevalence and overall behavioural composition (RQ1)}
RQ1 examined which coded history taking behaviours distinguish high-rated from low-rated encounters. For individual behaviours, we aggregated each learner's raw code counts within each week and compared the high-rated and low-rated groups using two-sided Mann-Whitney $U$ tests, while controlling for the false discovery rate (FDR) across codes within each week with the Benjamini-Hochberg (BH) procedure \citep{benjamini1995controlling}. We reported rank-biserial correlations for the Mann-Whitney $U$ tests, calculated as $r_{rb}=2U/(n_{\mathrm{high}}n_{\mathrm{low}})-1$, where positive values indicate higher counts in the high-rated group. Because raw counts are sensitive to dialogue length, we also checked total learner turns and repeated the code-level comparisons using code rates, computed as each code count divided by the learner's total number of coded turns in that week. In addition to testing each code separately, RQ1 examined whether the overall behavioural composition of a consultation differed by performance group. For this analysis, each learner dialogue was represented as an 11-dimensional behavioural profile, where each dimension was the count of one retained code. This profile-level analysis was included because two consultations may not differ strongly on a single code but may differ in the overall combination of routine questioning, symptom specification, checking, organisation, communication, and summarising. We computed Bray-Curtis dissimilarities from each learner's raw code count vector and ran PERMANOVA with 9{,}999 permutations per week \citep{anderson2001new}, reporting $R^2$ as the profile-level effect size. Rate-profile PERMANOVA used the same $R^2$ effect-size statistic and served as a length-control check.

Finally, because the main RQ1 analyses used weekly median splits to create performance groups, we conducted a continuous-score sensitivity analysis. The purpose was to examine whether the behavioural patterns observed in the high--low contrasts were consistent with score-related associations when the full range of weekly HT-Total scores was retained. For each retained code, we fitted a separate mixed-effects regression model with continuous HT-Total score as the outcome. The predictor of interest was the learner's rate for that code in that week, calculated as the code count divided by the learner's total number of coded turns. Total coded learner turns was included as a covariate to account for consultation length, and week was included as a fixed effect. Learner ID was included as a random intercept to account for repeated observations from the same learner across weeks. The model was:

\[
HTTotal_{ij} = \beta_0 + \beta_1 CodeRate_{ij} + \beta_2 TotalTurns_{ij} + \gamma_{Week_j} + u_i + \epsilon_{ij},
\]

where $i$ indexes learners and $j$ indexes weekly consultations. The same model was fitted separately for each of the 11 retained codes. Model estimates, standard errors, confidence intervals, unadjusted $p$ values, BH-adjusted $q$ values, and model fit indices are reported in Supplementary Table~S13.

\subsubsection{Local co-occurrence of coded history taking behaviours (RQ2)}
RQ2 examined whether history taking behaviours were combined differently in nearby turns in high- and low-rated encounters. We used Epistemic Network Analysis (ENA) to model local co-occurrence among the 11 retained utterance states after filtering out OS (i.e., SS, LO, RQ, PQ, RR, SI, SR, CK, FQ, RT, and CC were included) \citep{shaffer2016tutorial,csanadi2018coding}. ENA was built separately for each week, with each learner dialogue treated as one analytic unit. In ENA, the moving window defines the local context within which coded behaviours are treated as meaningfully connected, and this window should reflect the temporal grain of the process being studied \citep{shaffer2016tutorial,csanadi2018coding}. We therefore defined local co-occurrences using a backward moving window of four coded learner turns and accumulated these co-occurrences into one network representation for each dialogue. This a priori choice was intended to represent short history taking episodes rather than single next-step dependencies or whole-topic segments. Immediate next-state dependencies were analysed separately with TNA, whereas a much wider ENA window would make local co-occurrence less distinguishable from broader topic coverage across the consultation \citep{saint2020combining,saqr2024sequence}. As a sensitivity check, we repeated the ENA projection and edge-difference analyses with backward windows of three, five, and six turns. High- and low-rated encounters were first compared visually using weekly ENA difference networks. For statistical comparisons, each  dialogue's score on the first ENA projection dimension (MR1) was compared between groups using a two-sided Mann-Whitney $U$ test. Effect sizes for the MR1 comparisons were reported as rank-based $r$, calculated as the standardised Mann-Whitney test statistic divided by $\sqrt{N}$.

\subsubsection{Sequential transitions between coded history taking behaviours (RQ3)}
RQ3 examined how high- and low-rated encounters moved from one history taking behaviour to another. We used Transition Network Analysis (TNA) to model each dialogue as a first-order sequence over the 11 retained learner behavioural states and estimated conditional transition probabilities separately for each week and group \citep{saqr2024sequence}. For each transition, the effect estimate was the raw conditional probability difference, $\Delta p = P(\text{high-rated}) - P(\text{low-rated})$. This value can be interpreted as an unstandardised transition-level effect size metric in probability-point units; positive values indicate transitions more likely in high-rated encounters, and negative values indicate transitions more likely in low-rated encounters. Edge-level permutation tests evaluated whether the observed $\Delta p$ for each transition was larger than expected under permuted group labels. The interpretation focused on transitions that were statistically supported and educationally interpretable in relation to the process functions discussed in Section~\ref{sec:hist_regulation}: information gathering and symptom specification (RQ, SS), logical organisation (LO), cue-responsive follow-up and mechanism-oriented questioning (RR, PQ), checking and clarification (CK), summarising or restating information (SI, SR), facilitative communication (CC), and less productive questioning patterns such as repeated or vague questioning (RT, FQ). Full transition-level results for all tested source--target pairs are reported in Supplementary Table~S11.

\begin{table}[t]
\centering
\caption{Descriptive statistics for total coded learner turns by week and performance group.}
\label{tab:length_desc}
\small
\begin{tabular}{llrrrr}
\toprule
Week & Group & $n$ & Mean & SD & Median (IQR) \\
\midrule
W1 & High-rated & 109 & 103.01 & 28.45 & 98.00 (85.00--116.00) \\
W1 & Low-rated  & 97  & 77.40  & 28.62 & 73.00 (60.00--84.00) \\
W2 & High-rated & 106 & 70.62  & 10.19 & 71.00 (63.00--77.75) \\
W2 & Low-rated  & 101 & 65.17  & 11.39 & 66.00 (61.00--73.00) \\
W3 & High-rated & 107 & 73.91  & 10.16 & 75.00 (66.00--80.00) \\
W3 & Low-rated  & 98  & 65.48  & 9.19  & 65.00 (60.00--72.00) \\
W4 & High-rated & 104 & 75.90  & 12.00 & 75.00 (69.00--81.00) \\
W4 & Low-rated  & 102 & 66.59  & 11.30 & 67.00 (61.25--73.00) \\
W5 & High-rated & 105 & 82.51  & 18.02 & 80.00 (73.00--87.00) \\
W5 & Low-rated  & 101 & 68.72  & 11.81 & 69.00 (61.00--76.00) \\
\bottomrule
\end{tabular}
\end{table}

%% file: sections/04-results.tex
\section{Results}\label{sec:results}
We report results in the order of the three research questions: behavioural prevalence and overall behavioural composition (RQ1), local co-occurrence structure (RQ2), and sequential transitions (RQ3). Because performance groups were defined separately within each week, the results are interpreted as within-week contrasts.

\subsection{Behavioural prevalence and overall behavioural composition (RQ1)}\label{sec:results_rq1}

RQ1 examined consultation length, individual coded history taking behaviours, and overall behavioural composition. Table~\ref{tab:length_desc} reports total coded learner turns by week and performance group. Descriptive statistics for the 11 retained code counts are reported in Supplementary Table~S3, and descriptive statistics for code rates are reported in Supplementary Table~S4.

High-rated consultations contained more coded learner turns than low-rated consultations in every week. The descriptive gap was largest in W1 and smallest in W2. Full consultation-length tests, including Mann--Whitney $U$, unadjusted $p$ values, BH-adjusted $q$ values, and rank-biserial correlations, are reported in Supplementary Table~S2. Mann--Whitney $U$ tests are reported with group sizes rather than degrees of freedom.

\begin{table}[t]
\centering
\caption{Summary of count-based Mann--Whitney $U$ test results across weeks after BH within each week across codes. Reported are the number of weeks with significant group differences ($q<.05$), the significant weeks, the mean BH-adjusted $q$ value, and the mean effect size as rank-biserial correlation ($r_{rb}$) across those significant weeks. Full descriptive statistics are reported in Supplementary Tables~S3--S5.}
\label{tab:mannwhitney}
\small \setlength{\tabcolsep}{3pt} \renewcommand{\arraystretch}{1.08} \begin{tabularx}{\linewidth}{@{} >{\raggedright\arraybackslash}X >{\centering\arraybackslash}p{0.12\linewidth} >{\raggedright\arraybackslash}p{0.23\linewidth} >{\raggedleft\arraybackslash}p{0.12\linewidth} >{\raggedleft\arraybackslash}p{0.10\linewidth} @{}} \toprule Code (Name) & \shortstack{Sig.\\weeks\\($q<.05$)} & Weeks significant & \shortstack{Mean $q$\\(BH)} & \shortstack{Mean\\$r_{rb}$} \\
\midrule
SI (Summarising \& Integrating)   & 5 & W1, W2, W3, W4, W5 & 0.01158 & 0.225 \\
RQ (Routine Question)             & 5 & W1, W2, W3, W4, W5 & 0.02044 & 0.308 \\
LO (Logical Organisation)         & 4 & W1, W3, W4, W5     & $9.32\times10^{-6}$ & 0.424 \\
CC (Facilitative Communication)   & 4 & W1, W2, W4, W5     & 0.00434 & 0.295 \\
SS (Specifying Symptoms)          & 4 & W1, W3, W4, W5     & 0.01055 & 0.320 \\
FQ (Fuzzy Question)               & 2 & W1, W4             & 0.02341 & 0.174 \\
CK (Checking)                     & 1 & W1                 & $4.50\times10^{-5}$ & 0.332 \\
RR (Relevant Response)            & 1 & W5                 & 0.04057 & 0.164 \\
\bottomrule
\end{tabularx}
\end{table}

The raw-count comparisons showed higher counts for several coded behaviours in high-rated consultations (Table~\ref{tab:mannwhitney}). SI and RQ differed in all five weeks. LO, CC, and SS each differed in four weeks. FQ, CK, and RR showed more limited differences, and RT, SR, and PQ did not show reliable raw-count differences after BH correction. Across significant contrasts, mean rank-biserial correlations ranged from 0.164 for RR to 0.424 for LO.

After normalising counts by total coded learner turns, fewer differences remained. Significant rate-based contrasts were found for LO in W3 and W5, SI in W1 and W3, CK in W1, and RQ in W3. The rate-based results therefore narrow the raw-count interpretation: high-rated consultations were longer and contained more coded behaviours, but proportional differences were concentrated in a smaller set of behaviours, especially organisation, summarising/integrating, and checking.

The overall behavioural composition also differed between groups. Raw-count PERMANOVA was significant in all five weeks, with effect sizes ranging from $R^2=.0225$ in W2 to $R^2=.1306$ in W1. Rate-profile PERMANOVA showed significant composition differences in W1 ($R^2=.012$, $p=.041$) and W4 ($R^2=.013$, $p=.016$), while W2, W3, and W5 were not significant and had small effect sizes ($R^2=.006$--$.009$). Complete raw-count and rate-profile PERMANOVA outputs, including degrees of freedom, sums of squares, mean squares, pseudo-$F$, permutation $p$ values, $R^2$, and number of permutations, are reported in Supplementary Tables~S6 and S7.

The continuous-score sensitivity analysis partly supported the median-split findings. In separate mixed-effects models using continuous HT-Total as the outcome and controlling for week and total coded learner turns, LO ($\beta=1.076$, $p<.001$), SI ($\beta=1.269$, $p<.001$), and CC ($\beta=.616$, $p=.016$) were positively associated with HT-Total. SS was negatively associated with HT-Total ($\beta=-.615$, $p=.024$), and RQ was not statistically reliable. Full model estimates, standard errors, confidence intervals, unadjusted $p$ values, BH-adjusted $q$ values, and model fit indices are reported in Supplementary Table~S13. The SS result requires particular caution. SS appeared more often in high-rated consultations in raw counts, but its rate was negatively associated with continuous HT-Total after controlling for week and total coded learner turns. This suggests a length/composition distinction: high-rated consultations may include more symptom-specific questions in absolute terms because they contain more learner turns overall, but a higher proportion of SS may indicate that the consultation remains focused on symptom detailing rather than moving toward organisation, checking, or synthesis.

\begin{figure}[t]
\centering
\includegraphics[width=\linewidth]{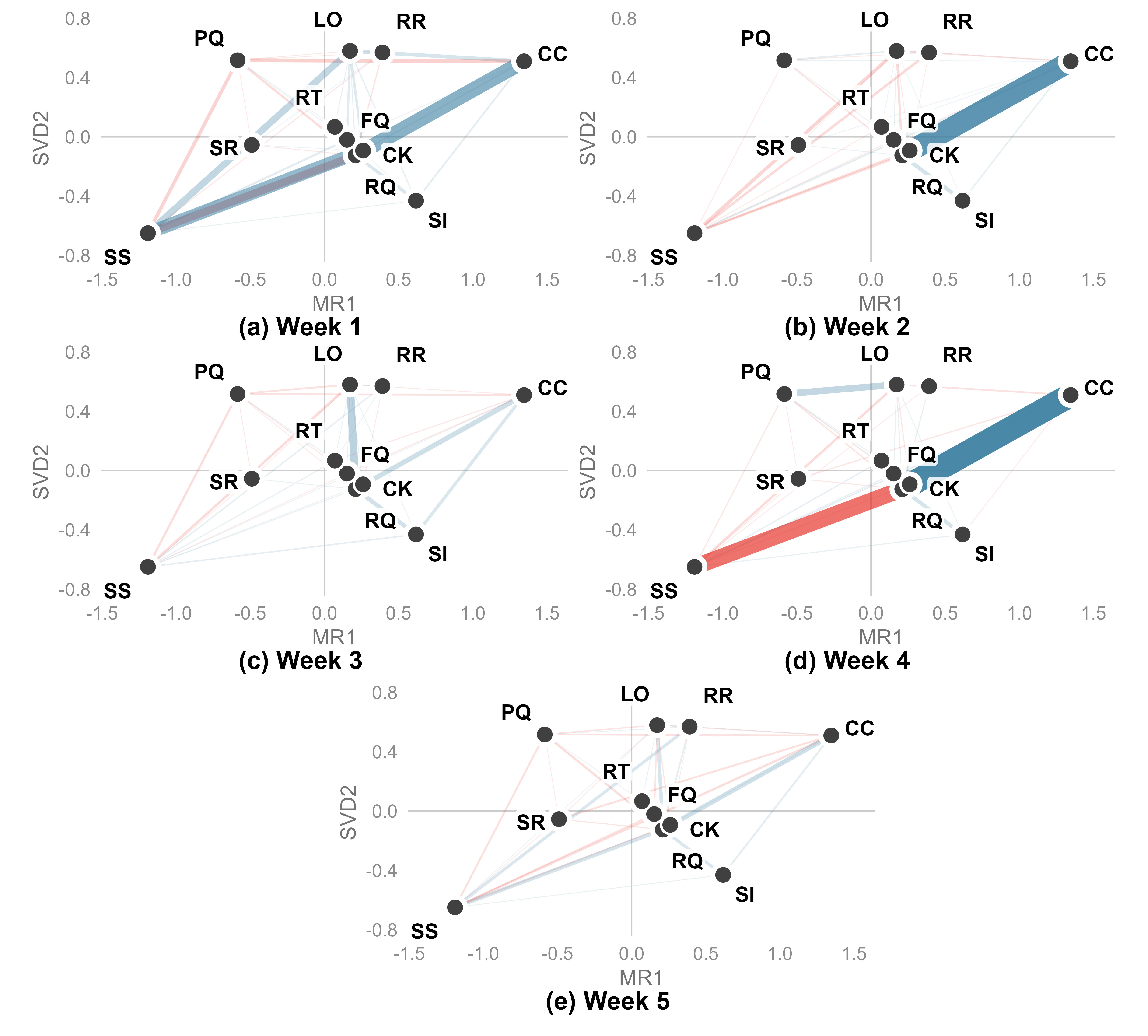}
\caption{Weekly ENA difference networks (high-rated minus low-rated). Nodes are coded learner behaviours from Table~\ref{tab:codebook}. Blue edges are stronger in the high-rated group, red edges are stronger in the low-rated group, and thicker edges indicate larger group differences. Layout and edge scaling are fixed across weeks; individual dialogue points and centroid confidence boxes are omitted to focus on co-occurrence contrasts.}
\label{fig:ena_diff_byweek}
\end{figure}

\subsection{Performance-related differences in co-occurrence structure (RQ2)}\label{sec:results_rq2}

RQ2 examined how coded history taking behaviours were placed near one another. Figure~\ref{fig:ena_diff_byweek} maps local pairings by week: blue edges mark pairings stronger in high-rated consultation sessions, red edges mark pairings stronger in low-rated consultation sessions, and thicker edges indicate larger group differences. The high-rated group scored significantly higher on the first ENA dimension (MR1) across all five weeks, indicating consistent group separation in local coordination. Mann-Whitney $U$ tests do not have degrees of freedom, so the tests are reported with weekly group sizes and rank-based effect sizes: W1, $n_{\mathrm{high}}=109$, $n_{\mathrm{low}}=97$, $U=6847.0$, $p<.001$, $r=.255$; W2, $n_{\mathrm{high}}=106$, $n_{\mathrm{low}}=101$, $U=6526.0$, $p=.006$, $r=.189$; W3, $n_{\mathrm{high}}=107$, $n_{\mathrm{low}}=98$, $U=6765.0$, $p<.001$, $r=.251$; W4, $n_{\mathrm{high}}=104$, $n_{\mathrm{low}}=101$, $U=6645.0$, $p=.001$, $r=.229$; and W5, $n_{\mathrm{high}}=105$, $n_{\mathrm{low}}=101$, $U=7057.0$, $p<.001$, $r=.286$. The window-size sensitivity check showed the same MR1 group separation for windows of three, five, and six turns in every week (all $p<.007$). Edge-difference vectors from the alternative windows were also strongly aligned with the four-turn solution, with mean Pearson correlations of .970, .980, and .952 for windows of three, five, and six turns, respectively. The clearest pattern concerned the placement of RQ. In W1, W2, and W4, the RQ-CC edge was stronger in high-rated consultation sessions. This means that routine inquiry was more often placed near moves that oriented the patient, signalled a topic shift, or maintained the interaction, rather than appearing only as a checklist-like progression. In W3, RQ was more tightly connected with LO and SI, indicating that routine inquiry was locally tied to organising the case and pulling information together. In W4, stronger edges around SI again placed synthesis close to ongoing inquiry. SS showed a more case-sensitive pattern. The SS-RQ pairing was stronger in low-rated consultation sessions in W1 and W4, where local episodes were more dominated by symptom probing and routine inquiry, but it was stronger in high-rated consultations in W5. This suggests that probing symptom detail was not uniformly associated with high performance; its meaning depended on the surrounding moves and the weekly case. Across the five tasks, the co-occurrence evidence shows that high-rated consultations differed most in how RQ, CC, LO, and SI were locally connected, while SS became useful when it was embedded in a broader inquiry structure rather than simply paired mainly with further routine questioning.

\subsection{Temporal sequencing and transition differences (RQ3)}\label{sec:results_rq3}

\begin{figure}[t]
\centering
\begin{subfigure}{0.32\textwidth}\includegraphics[width=\linewidth]{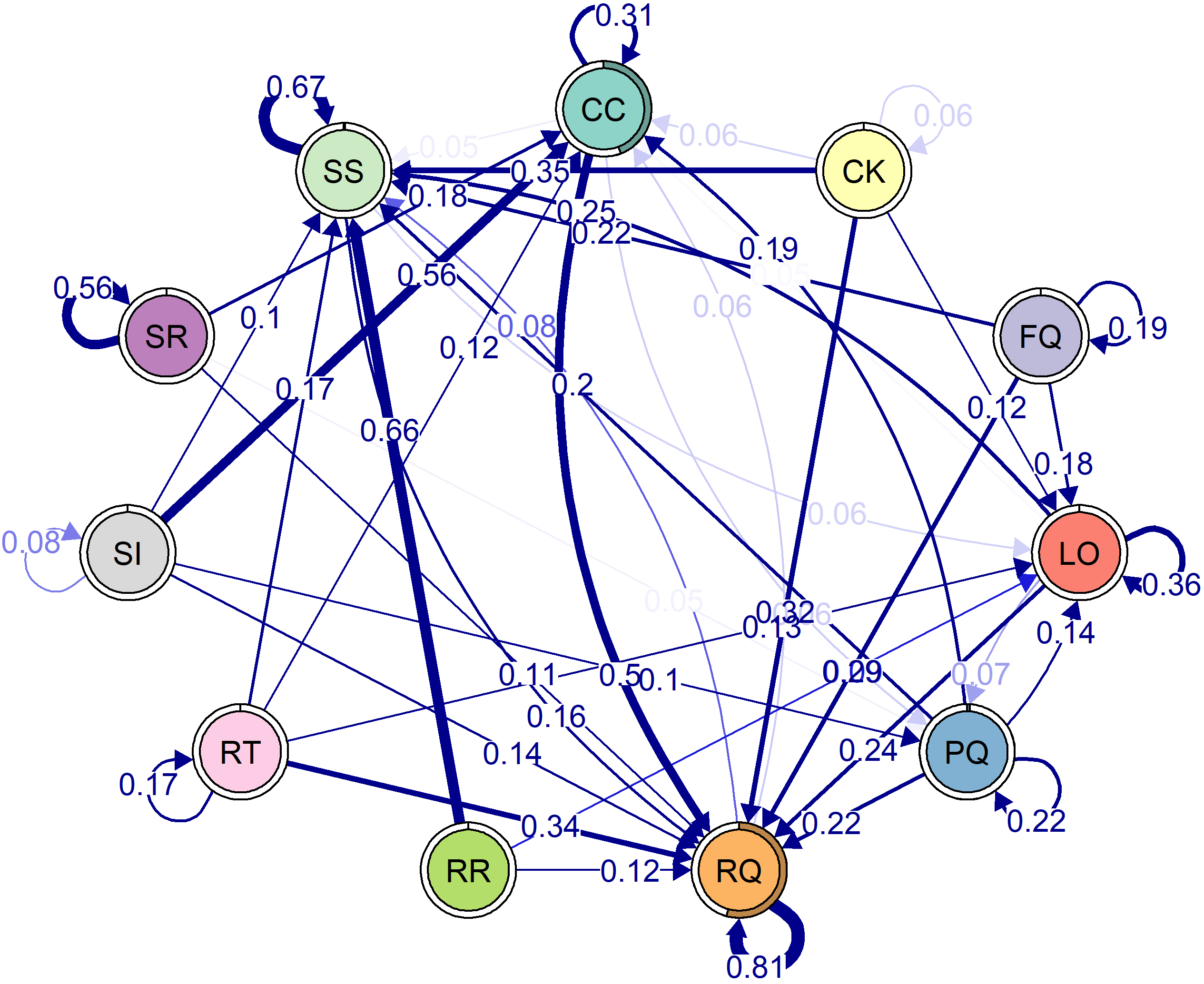}\caption{W1}\end{subfigure}
\begin{subfigure}{0.32\textwidth}\includegraphics[width=\linewidth]{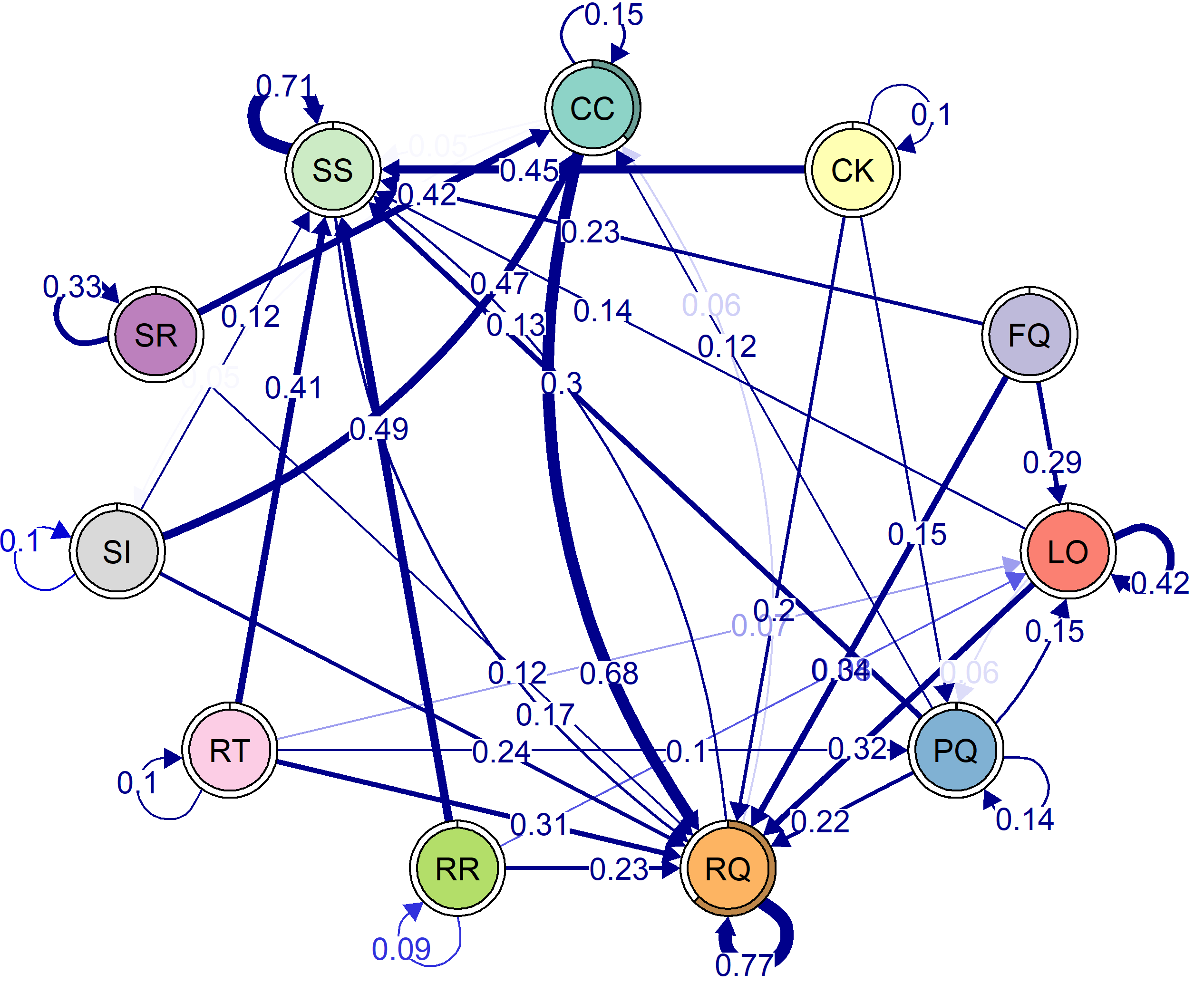}\caption{W2}\end{subfigure}
\begin{subfigure}{0.32\textwidth}\includegraphics[width=\linewidth]{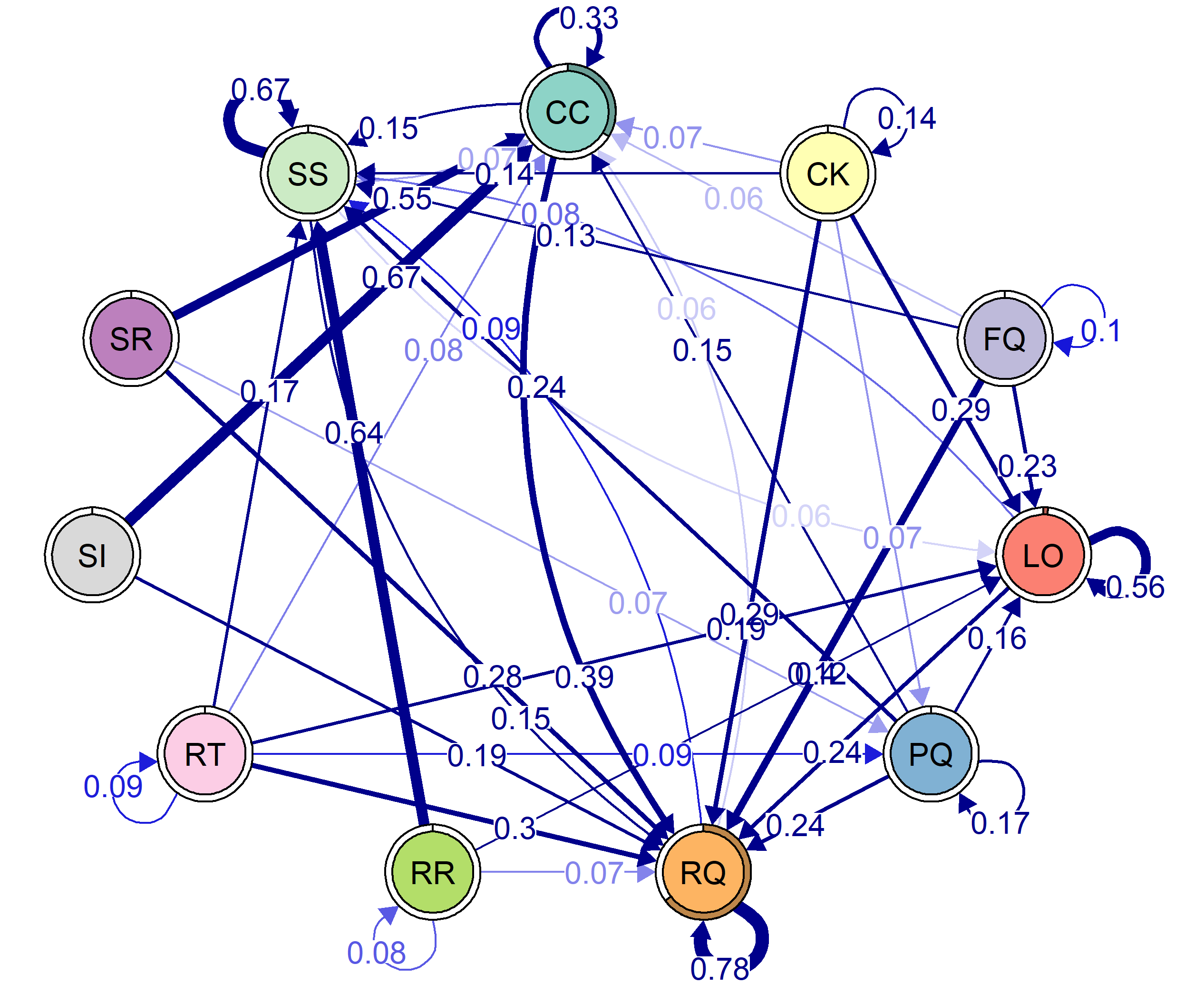}\caption{W3}\end{subfigure}

\medskip
\begin{subfigure}{0.32\textwidth}\includegraphics[width=\linewidth]{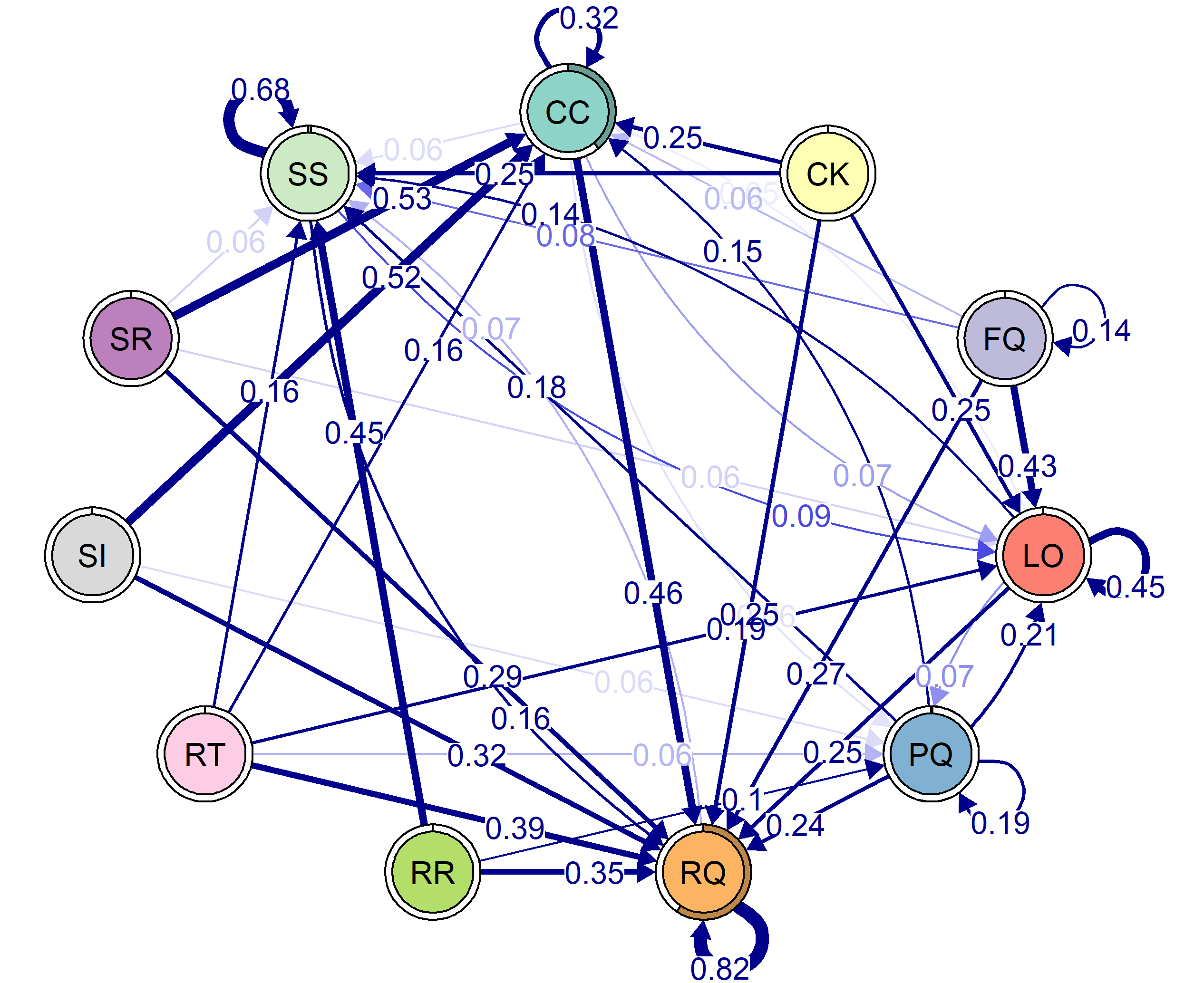}\caption{W4}\end{subfigure}
\begin{subfigure}{0.32\textwidth}\includegraphics[width=\linewidth]{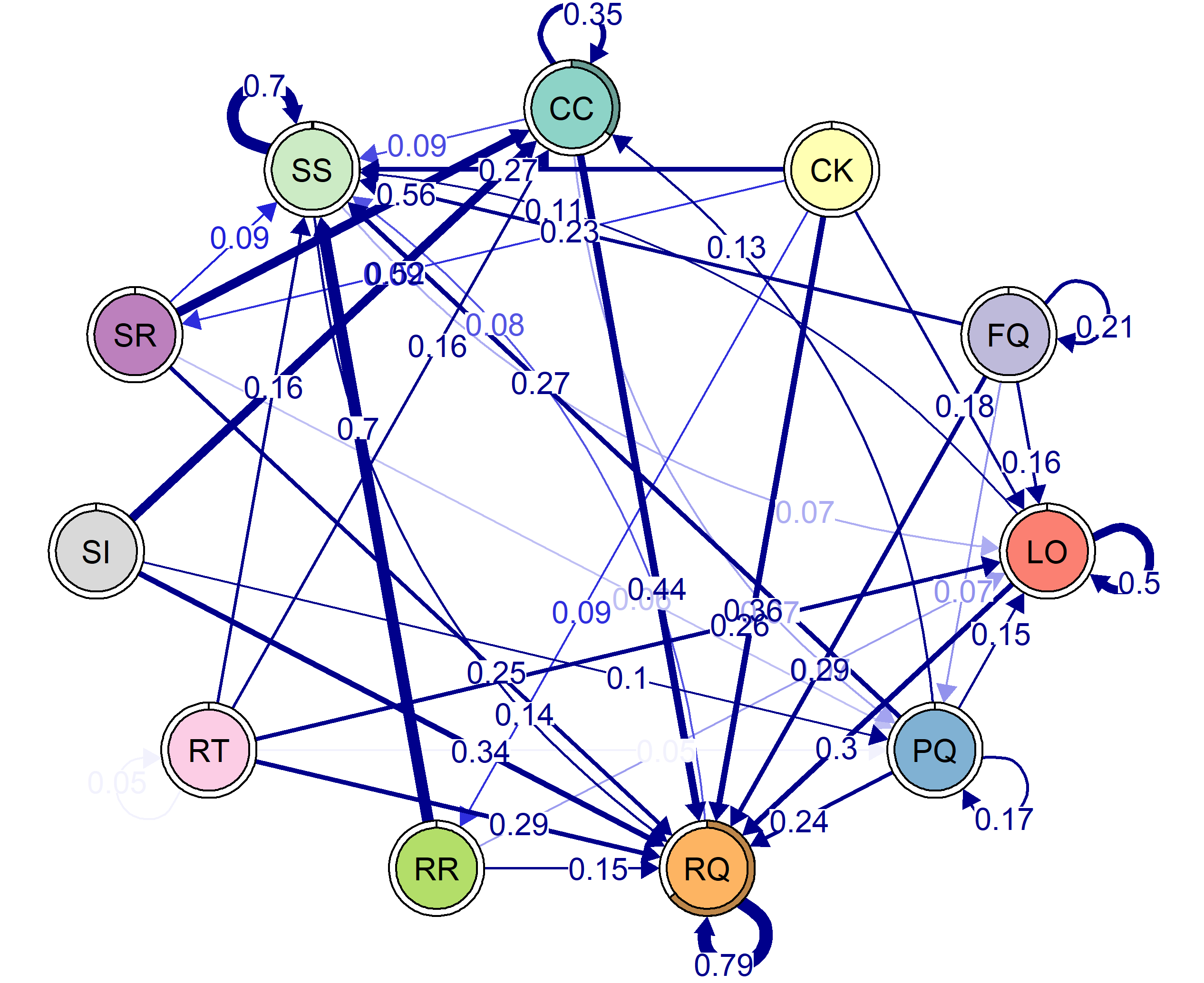}\caption{W5}\end{subfigure}
\begin{subfigure}{0.32\textwidth} 
\end{subfigure}

\caption{Full-cohort weekly transition networks. Nodes are coded learner behaviours from Table~\ref{tab:codebook}; directed edges show common next-state transitions in the full cohort. Self loops indicate repeated use of the same behavioural state.}
\label{fig:tna_compare_byweek}
\end{figure}

\begin{figure}[t]
\centering
\begin{subfigure}{0.32\textwidth}\includegraphics[width=\linewidth]{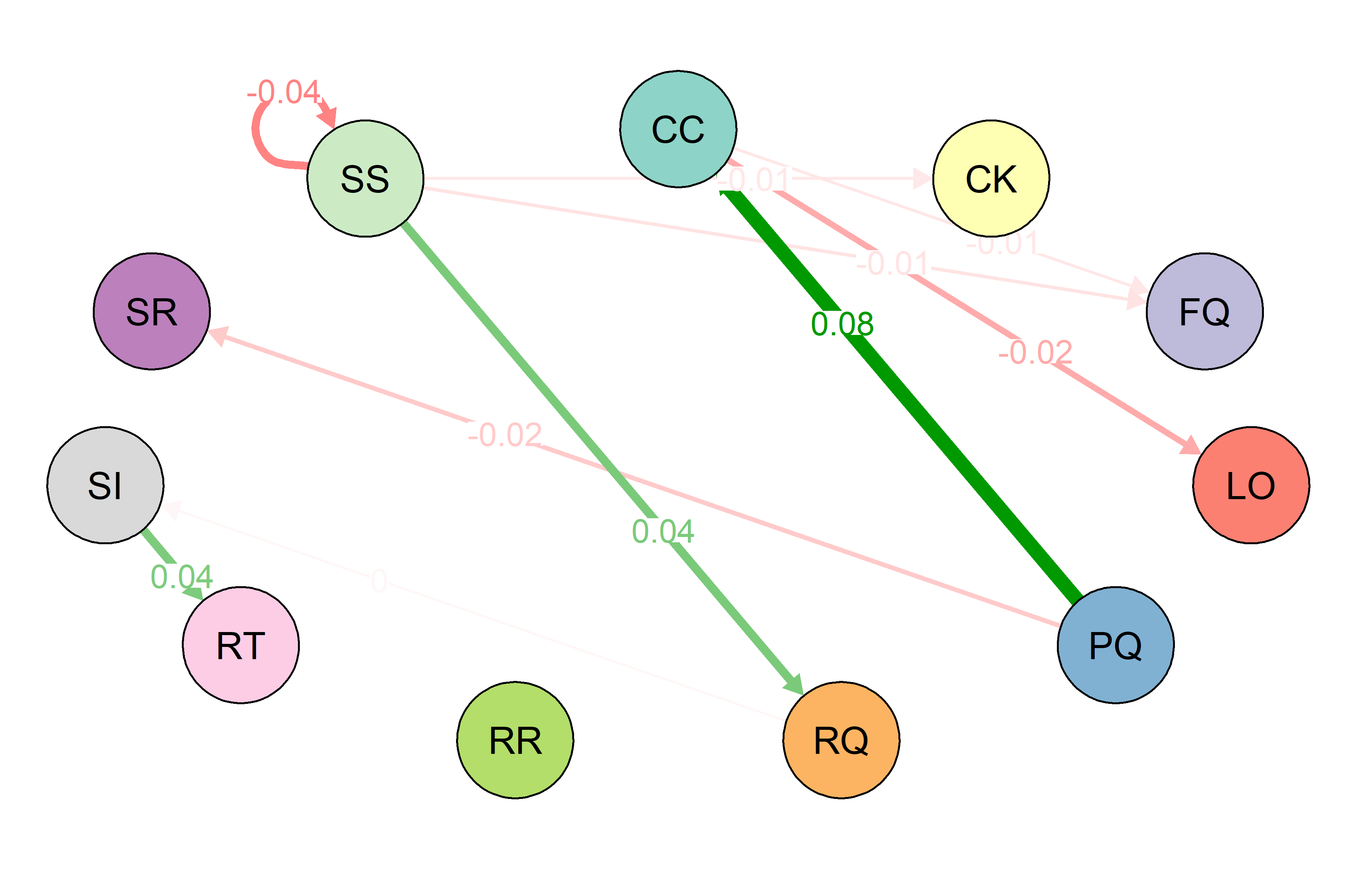}\caption{W1}\end{subfigure}
\begin{subfigure}{0.32\textwidth}\includegraphics[width=\linewidth]{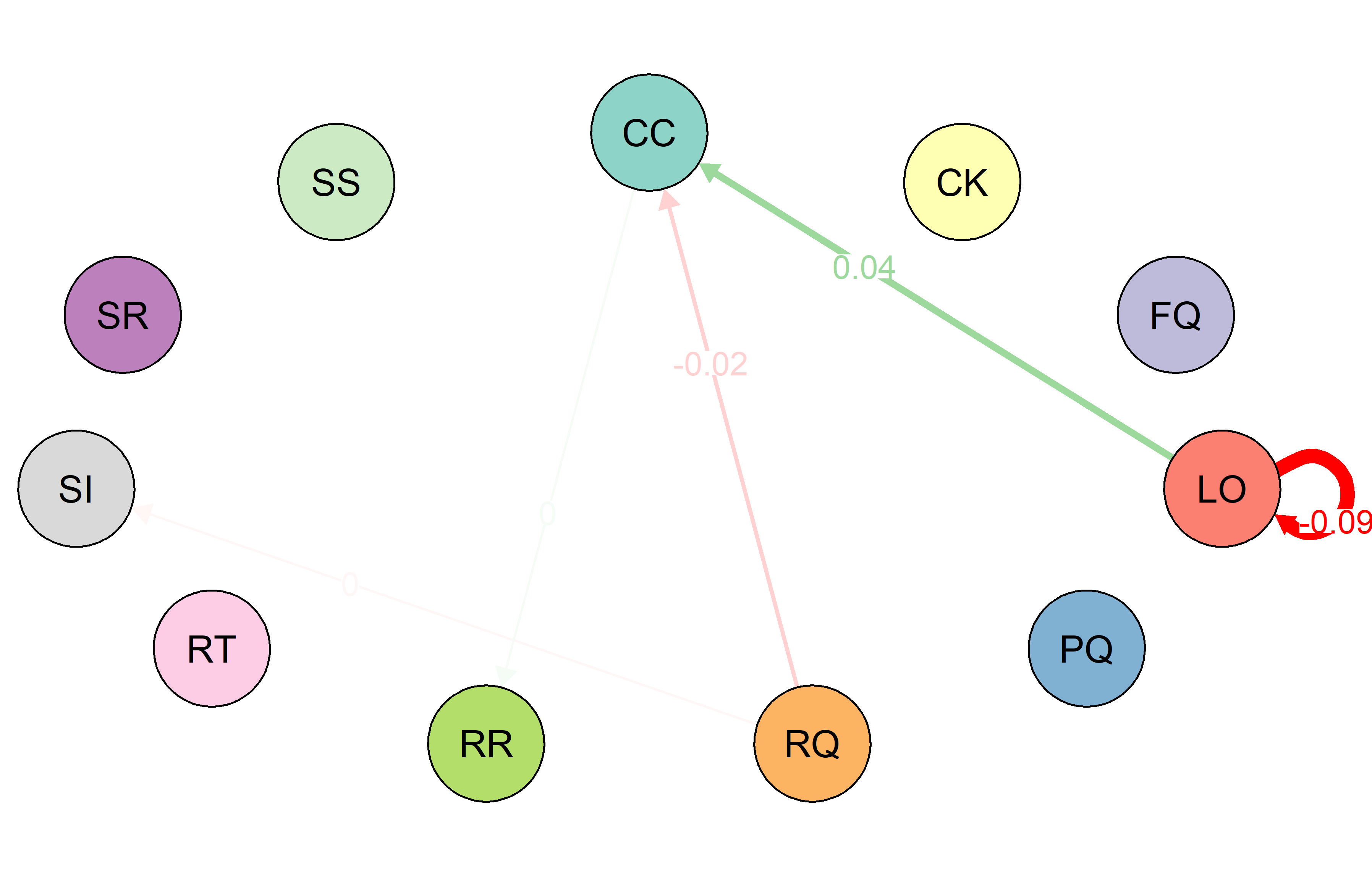}\caption{W2}\end{subfigure}
\begin{subfigure}{0.32\textwidth}\includegraphics[width=\linewidth]{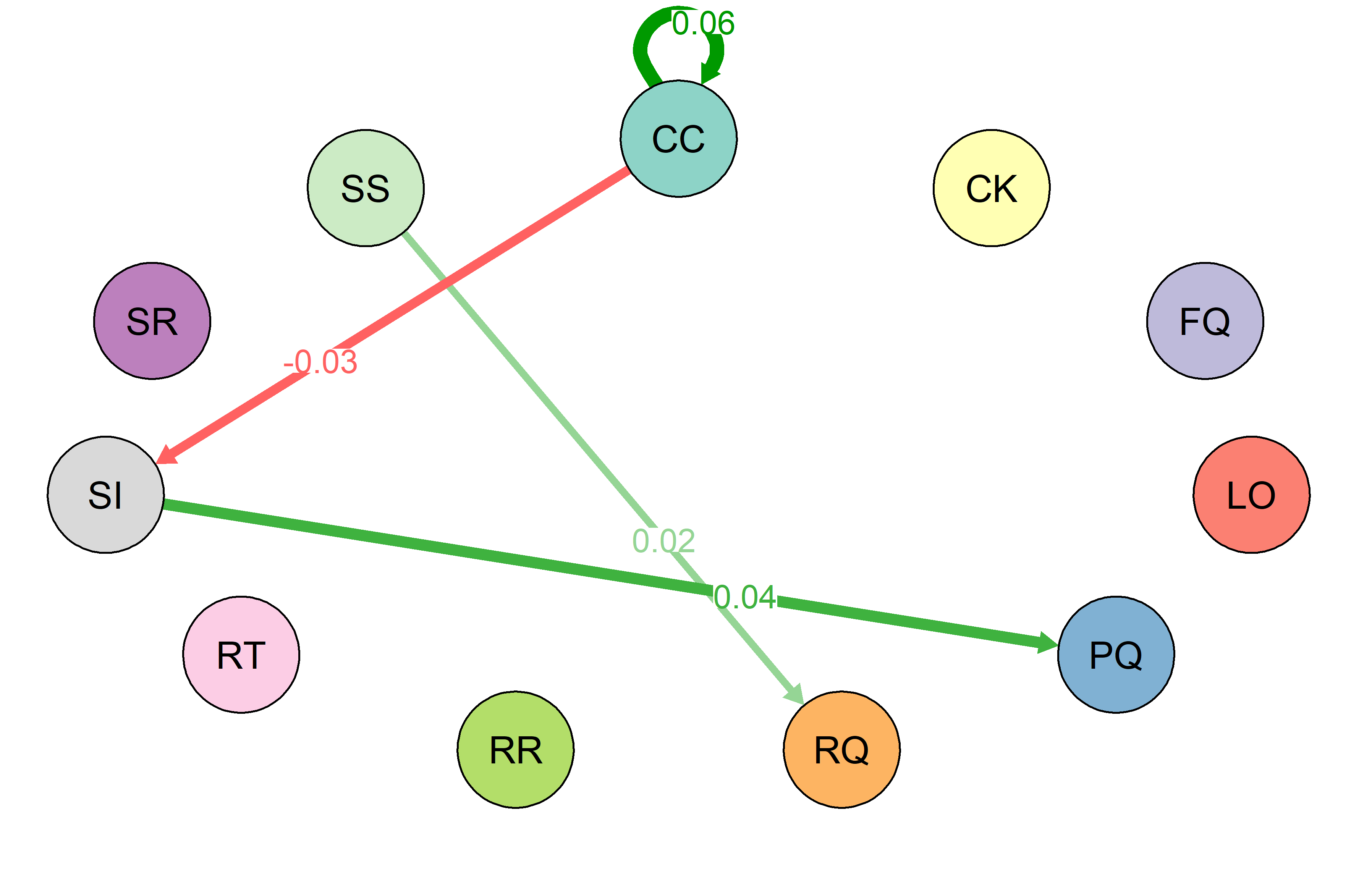}\caption{W3}\end{subfigure}

\medskip

\begin{subfigure}{0.32\textwidth}\includegraphics[width=\linewidth]{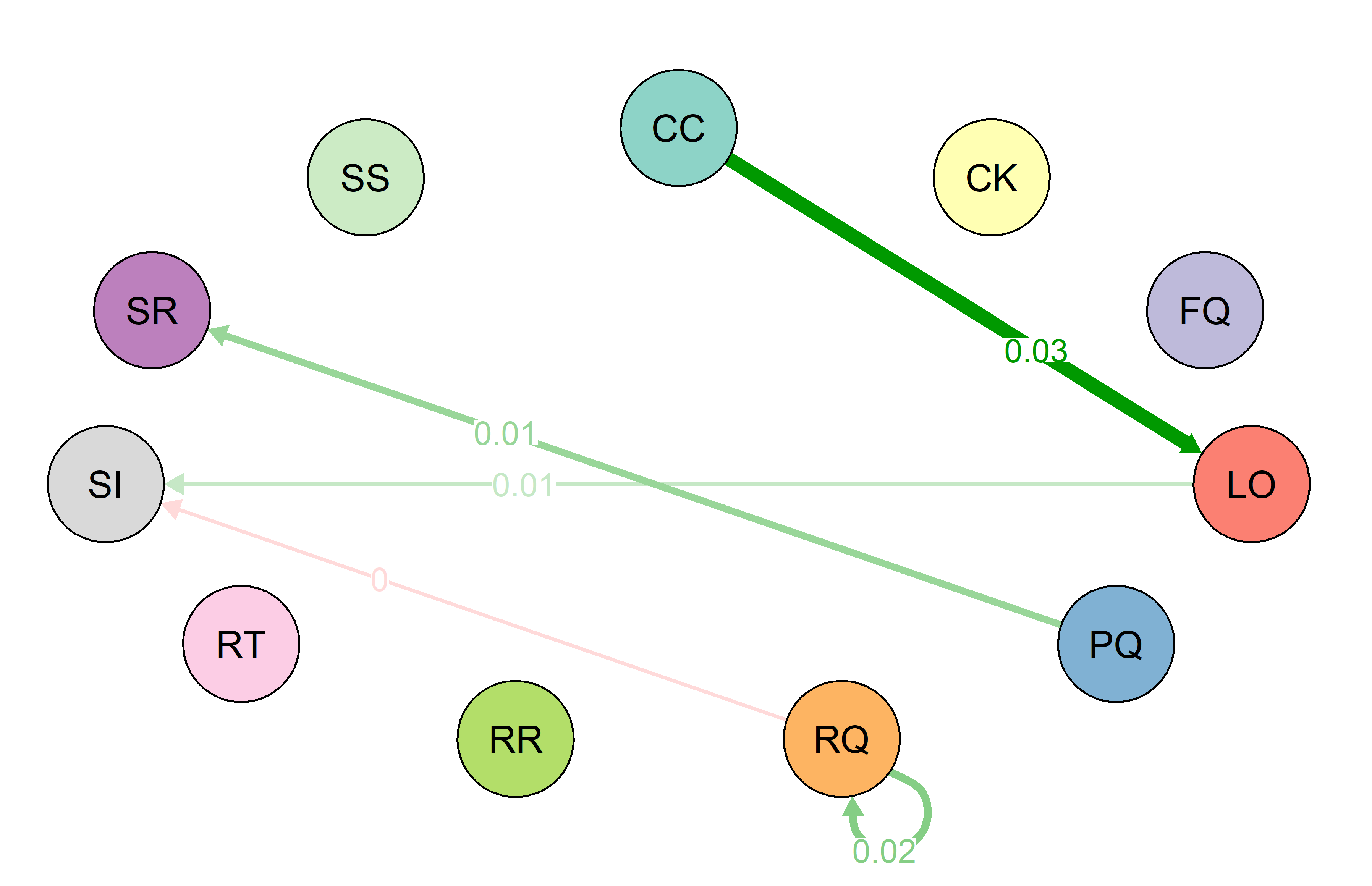}\caption{W4}\end{subfigure}
\begin{subfigure}{0.32\textwidth}\includegraphics[width=\linewidth]{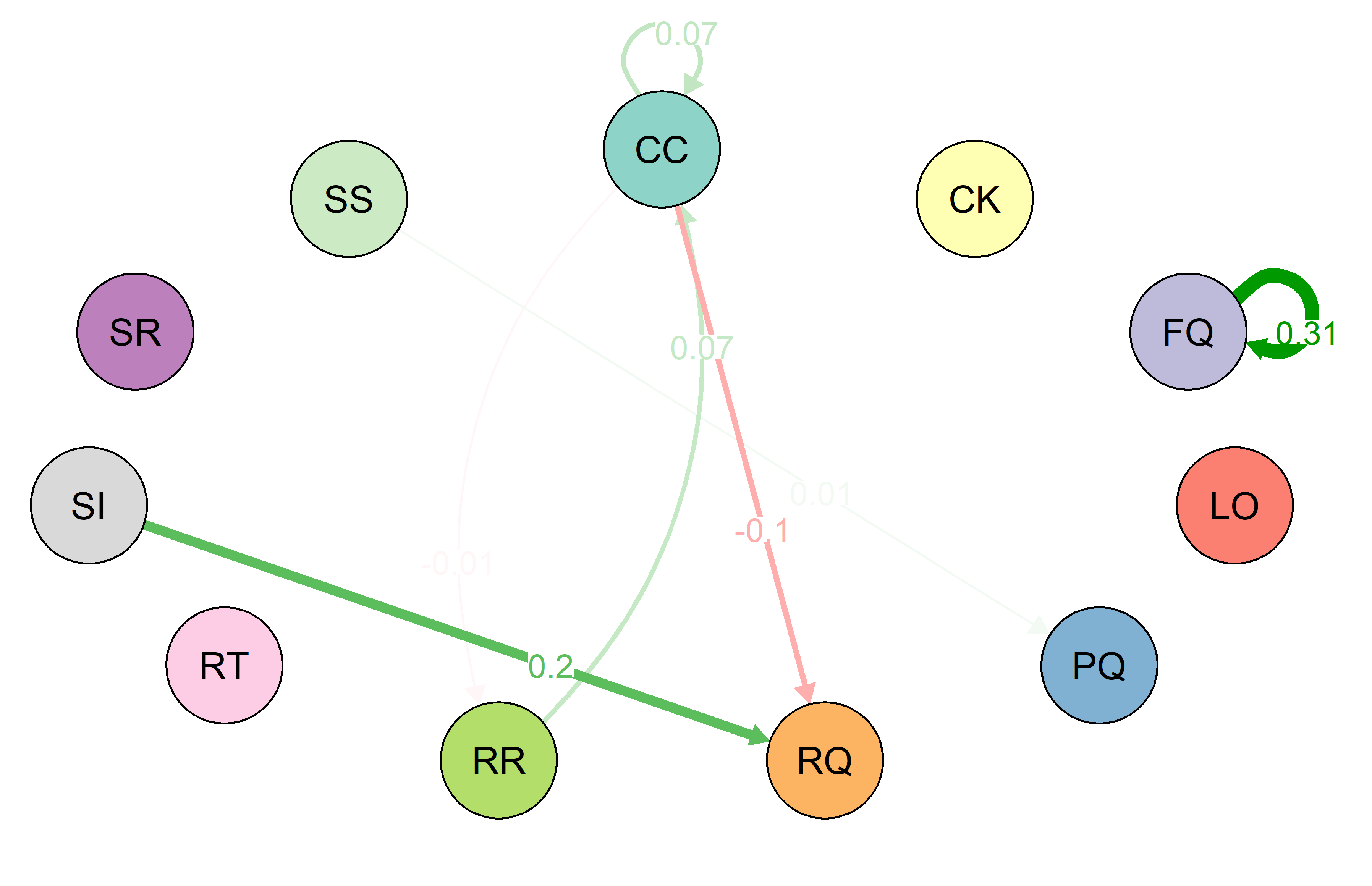}\caption{W5}\end{subfigure}
\begin{subfigure}{0.32\textwidth} 
\end{subfigure}

\caption{Transition differences by week based on edge-level permutation tests ($p<.05$). Nodes represent coded learner behaviours from Table~\ref{tab:codebook}. Directed edges are labelled by $\Delta p = P(\text{high}) - P(\text{low})$, the raw conditional-probability difference estimate. Green edges indicate transitions more likely in the high-rated group, red edges indicate transitions more likely in the low-rated group, and thicker edges indicate larger absolute $\Delta p$ values.}
\label{fig:tna_sig_byweek}
\end{figure}

RQ3 examined which history taking behaviour usually followed a given history taking behaviour. The analysis separates the shared sequence structure of the weekly consultations from the points where high- and low-rated consultations diverged within the shared sequence structure. Figure~\ref{fig:tna_compare_byweek} shows the full-cohort transition routine for each week, and Figure~\ref{fig:tna_sig_byweek} marks transition contrasts within the weekly transition routine. Positive $\Delta p$ values indicate transitions more likely in high-rated encounters; negative values indicate transitions more likely in low-rated encounters. The reported $\Delta p$ values are raw conditional-probability difference estimates and can be read as unstandardised transition-level effect-size metrics.

The full-cohort networks show that all five weekly tasks shared a similar transition routine. The most visible structure was repeated movement within RQ and SS, with additional repeated use of LO and CC. Most learners, regardless of performance group, spent substantial portions of the encounter in routine inquiry, symptom detail work, organisation, or managing the conversation.

The group contrasts were differences inside a shared routine rather than completely different sequence structures. CC appeared in different parts of the shared routine. In W2, high-rated encounters more often moved from LO to CC (LO$\rightarrow$CC; effect estimate $\Delta p=.039$, $p=.002$), whereas low-rated consultations more often moved from RQ to CC (RQ$\rightarrow$CC; effect estimate $\Delta p=-.018$, $p=.008$). In W5, high-rated consultations more often moved from RR to CC (RR$\rightarrow$CC; effect estimate $\Delta p=.079$, $p=.001$), whereas low-rated consultations more often moved from CC back to RQ (CC$\rightarrow$RQ; effect estimate $\Delta p=-.092$, $p=.006$). CC tended to follow organising the case or following up patient cues in high-rated consultations, but it more often returned the learner to routine questioning in low-rated consultations. The second difference concerns what happened after SI. In W3, high-rated consultations more often moved from SI to PQ (SI$\rightarrow$PQ; effect estimate $\Delta p=.045$, $p=.001$), and in W4 they more often moved from SI to CK (SI$\rightarrow$CK; effect estimate $\Delta p=.030$, $p=.004$). By contrast, W1 low-rated consultations more often moved from SI to RT (SI$\rightarrow$RT; effect estimate $\Delta p=-.050$, $p=.001$). The same summary move could lead into mechanism-focused questioning or confirming details, or it could be followed by repeated questions. These transition patterns suggest that the same behaviour, especially SI or CC, carried different instructional meanings depending on what it led to next.

%% file: sections/05-discussion.tex
\section{Discussion}\label{sec:discussion}

\subsection{Performance-related process evidence in GenAI VP history taking}\label{sec:discussion_process_evidence}


Section~\ref{sec:la_process_evidence} identified a problem in using GenAI VP dialogue logs for clinical reasoning education: teacher-rated scores can anchor consultation quality, but they do not explain how the consultation unfolded. The RQ1 results addressed the first part of this problem by linking teacher-rated consultation quality to coded learner behaviours (Section~\ref{sec:results_rq1}). The clearest difference was behavioural volume. High-rated consultations contained more coded learner turns in every weekly case and higher raw counts of several behaviours. This finding should not be dismissed as a nuisance effect, because longer consultations may give learners more opportunity to gather information, organise the case, check uncertainty, and summarise. At the same time, the length-adjusted analyses narrowed the interpretation. Rate-based comparisons were more selective, rate-profile PERMANOVA effects were small, and the continuous-score sensitivity analysis identified LO, SI, and CC as the clearest positive score-related behaviours. The contribution of RQ1 is therefore not a claim that performance differences were independent of consultation length. Rather, RQ1 shows that raw behavioural volume was the most stable group difference, while length-adjusted and continuous-score analyses helped identify which behaviours remained most closely associated with assessed history-taking quality.

This distinction is especially important for interpreting symptom-specific questioning. In the raw-count analyses, SS appeared more often in high-rated consultations. However, in the continuous-score sensitivity analysis, SS was negatively associated with HT-Total after controlling for week and total coded learner turns. This is not merely an unstable result; it is a length--composition reversal. High-rated consultations may contain more symptom-specific questions in absolute terms because they contain more learner turns overall, but a higher proportion of SS may indicate that the consultation remains concentrated on symptom detailing rather than moving toward organisation, checking, or synthesis. SS should therefore not be interpreted as uniformly beneficial. Its educational meaning depends on how symptom details are used in the wider consultation.

The RQ2 results explain why the raw-count results should not be read as a simple volume effect. ENA showed that high-rated consultations more often connected routine questioning and symptom-specific questioning with communication, organisation, checking, and summarising (Section~\ref{sec:results_rq2}). This matters because RQ and SS are not inherently strong or weak behaviours. A routine question placed near LO, CK, or SI can contribute to an organised clinical account; a routine question placed mainly near further routine questioning can remain checklist-like. This finding extends communication and history taking assessment frameworks by showing that the educational meaning of a coded behaviour depends partly on its local coordination with other behaviours \citep{kurtz2003marrying,roter2002roter}.

The RQ3 results added temporal direction to this interpretation. TNA showed that SI had different meanings depending on what followed it (Section~\ref{sec:results_rq3}). SI followed by CK or PQ suggested that summarising helped the learner verify uncertainty or pursue mechanism-oriented follow-up. SI followed by RT suggested that the learner restated information without using the summary to guide the next question. CC also depended on sequence position: CC after LO or RR suggested that communication supported the flow of an organised consultation, whereas CC followed by RQ suggested a return to general information gathering. These findings show why sequence evidence is needed in addition to prevalence and co-occurrence evidence: temporal order helps distinguish behaviours that are merely present from behaviours that shape the next step of the consultation \citep{saint2022temporally,saqr2024sequence}.

The main contribution is therefore not that high-rated learners used a separate set of behaviours. Rather, high-rated consultations showed stronger coordination among ordinary history taking behaviours. This interpretation is consistent with a metacognition-informed view of history taking, in which learners monitor what has been established, identify uncertainty, organise information, and decide how later questions should develop the consultation \citep{winne2022modeling,rakovic2023harnessing}. The interpretation remains cautious because the data record learner utterances, not learners' conscious planning or monitoring. A turn coded as LO, SI, CK, RR, or PQ is consistent with organising, synthesising, monitoring, cue response, or hypothesis testing, but it does not directly measure those cognitive processes.

The patterns observed here are not specific to GenAI VPs. Structuring the consultation, checking patient information, summarising, and following clinically relevant cues are also emphasised in clinical communication and history taking assessment frameworks \citep{kurtz2003marrying,roter2002roter,haring2017observable,furstenberg2020assessing}. The same analytic approach could be applied to SP or OSCE encounters if those encounters were recorded, transcribed, segmented into learner turns, and coded with the same behavioural scheme. Prevalence analysis could compare how often learners used behaviours such as CK, SI, or SS; ENA could examine whether these behaviours were locally connected with LO, CC, or RQ; and TNA could examine whether summaries led to checking, mechanism-oriented questioning, or repetition. The practical difference is that GenAI VP systems already store complete turn-by-turn logs during routine practice, whereas SP and OSCE settings require additional recording, transcription, segmentation, and coding.

\subsection{Instructional uses of performance-linked dialogue patterns}\label{sec:instructional_uses}
The results suggest three uses for instruction and system design. First, teacher-facing reports could help instructors locate transcript segments worth reviewing. For example, a report could flag a long stretch of RQ or SS that is not followed by LO, CK, or SI. The teacher could then discuss with the learner what information had already been established, what remained uncertain, and how the next question could have been guided by the case information already collected.

Second, GenAI VP systems could use sequence patterns to support adaptive practice. The transition results suggest that SI is a useful point for intervention because it can either redirect the consultation or fail to change the subsequent questioning. If a learner summarises and then repeats an already answered question, the system could prompt the learner to clarify an uncertainty or ask a mechanism-oriented follow-up. If a learner summarises and then moves to CK or PQ, the system could reinforce the use of summaries as a bridge to verification or reasoning-oriented questioning.

Third, aggregated dialogue patterns could inform curriculum review. If many learners ask symptom-specific questions but rarely connect them to checking, organisation, or summarising, the issue may not be lack of symptom coverage. It may be difficulty using symptom information to structure the consultation. Similarly, if CC often leads back to routine questioning, learners may need more guidance on using communication to maintain consultation flow rather than restarting general data gathering.

These uses should not be treated as ready-made scoring rules. The present study identified performance-linked dialogue patterns; it did not test whether reports, prompts, or teacher review based on these patterns improve later history taking. Future studies should examine whether teachers find the patterns interpretable, whether learners can act on them, and whether instruction based on these patterns improves later GenAI VP, SP, or OSCE performance.

\subsection{Limitations and future work}

Several design features shape the interpretation of the findings. First, the performance anchor was the course-based history taking score. This anchor fits the teaching context, but future studies should test whether the same process signatures align with OSCE performance, expert panel ratings, diagnostic reasoning assessments, or later clinical interview performance \citep{misra2024osce}. The present analyses also focused on weekly performance contrasts rather than individual growth; longitudinal models are still needed to examine whether the same learners develop stronger coordination and transition patterns across repeated GenAI VP encounters \citep{molenaar2023measuring}.

Second, the coded sequence focused on learner utterances. GenAI VP responses were retained as interactional context but were not coded as behavioural states. This learner-only focus limits claims about the full learner--GenAI VP interaction, because a learner's next move may depend on the VP's cue, wording, ambiguity, or response quality \citep{jarvela2023human}. Coding the VP side would help distinguish learner uncertainty, missed patient cues, and limitations in the VP response. The study was also conducted in one course, one GenAI VP environment, and five chest pain cases generated with GPT-3.5. Replication with newer GenAI models, other symptoms, other institutions, and different VP designs is needed to determine which process signatures are stable features of coherent inquiry and which are case-, system-, or curriculum-specific.

Finally, dialogue logs were the sole source of process evidence. Without think-aloud protocols, retrospective interviews, eye-tracking, or other cognitive measures, the metacognitive interpretation remains an inference from behaviour rather than direct observation. Future work should combine GenAI VP logs with complementary process data and validated GenAI-assisted coding pipelines. Such work can test whether process-signature feedback is interpretable to teachers, actionable for learners, and effective in improving later history taking performance \citep{veenman2007assessment}.

%% file: sections/06-conclusion.tex
\section{Conclusion}\label{sec:conclusion}
The present study analysed GenAI VP history taking dialogues to examine whether open-ended consultation logs can be transformed into interpretable process evidence. By comparing high- and low-rated consultations through behavioural prevalence, ENA, and TNA, the study showed that stronger performance was not explained only by asking more questions or producing longer dialogues. Rather, high-rated consultations showed how learners wove basic history questions, symptom exploration, conversation management, clarification, case structuring, and interim synthesis into an organised clinical account, then used that account to verify details or pursue reasoning-oriented follow-up. These findings suggest that GenAI VP logs can make learners' history taking processes visible at scale, offering learning analytics a way to connect teacher-rated performance with concrete behavioural patterns that can inform future feedback and support for clinical reasoning practice.

%% file: sn-article.bbl
\begin{thebibliography}{}
\renewcommand{\doi}[1]{\url{https://doi.org/#1}}
\bibcommenthead

\bibitem [\protect \citeauthoryear {%
Anderson%
}{%
Anderson%
}{%
{\protect \APACyear {2001}}%
}]{%
anderson2001new}
\APACinsertmetastar {%
anderson2001new}%
\begin{APACrefauthors}%
Anderson, M.J.%
\end{APACrefauthors}%
\unskip\
\newblock
\APACrefYearMonthDay{2001}{}{}.
\newblock
{\BBOQ}\APACrefatitle {A new method for non-parametric multivariate analysis of variance} {A new method for non-parametric multivariate analysis of variance}.{\BBCQ}
\newblock
\APACjournalVolNumPages{Austral ecology}{26}{1}{32--46,}
\newblock

\newblock

\PrintBackRefs{\CurrentBib}

\bibitem [\protect \citeauthoryear {%
{Authors}%
}{%
{Authors}%
}{%
{\protect \APACyear {2026}}%
{\protect \APACexlab {{\protect \BCnt {1}}}}}]{%
chen2026developing}
\APACinsertmetastar {%
chen2026developing}%
\begin{APACrefauthors}%
{Authors}%
\end{APACrefauthors}%
\unskip\
\newblock
\APACrefYearMonthDay{2026{\protect \BCnt {1}}}{}{}.
\newblock
{\BBOQ}\APACrefatitle {Anonymized title} {Anonymized title}.{\BBCQ}
\newblock
\APACjournalVolNumPages{Anonymized venue}{}{}{,}
\newblock

\newblock

\PrintBackRefs{\CurrentBib}

\bibitem [\protect \citeauthoryear {%
{Authors}%
}{%
{Authors}%
}{%
{\protect \APACyear {2026}}%
{\protect \APACexlab {{\protect \BCnt {2}}}}}]{%
li2026flora}
\APACinsertmetastar {%
li2026flora}%
\begin{APACrefauthors}%
{Authors}%
\end{APACrefauthors}%
\unskip\
\newblock
\APACrefYearMonthDay{2026{\protect \BCnt {2}}}{}{}.
\newblock
{\BBOQ}\APACrefatitle {Anonymized title} {Anonymized title}.{\BBCQ}
\newblock
\APACjournalVolNumPages{Anonymized venue}{}{}{,}
\newblock

\newblock

\PrintBackRefs{\CurrentBib}

\bibitem [\protect \citeauthoryear {%
Benjamini%
\ \BBA {} Hochberg%
}{%
Benjamini%
\ \BBA {} Hochberg%
}{%
{\protect \APACyear {1995}}%
}]{%
benjamini1995controlling}
\APACinsertmetastar {%
benjamini1995controlling}%
\begin{APACrefauthors}%
Benjamini, Y.%
\BCBT {}\ \BBA {} Hochberg, Y.%
\end{APACrefauthors}%
\unskip\
\newblock
\APACrefYearMonthDay{1995}{}{}.
\newblock
{\BBOQ}\APACrefatitle {Controlling the false discovery rate: a practical and powerful approach to multiple testing} {Controlling the false discovery rate: a practical and powerful approach to multiple testing}.{\BBCQ}
\newblock
\APACjournalVolNumPages{Journal of the Royal statistical society: series B (Methodological)}{57}{1}{289--300,}
\newblock

\newblock

\PrintBackRefs{\CurrentBib}

\bibitem [\protect \citeauthoryear {%
Bennett%
}{%
Bennett%
}{%
{\protect \APACyear {2011}}%
}]{%
bennett2011formative}
\APACinsertmetastar {%
bennett2011formative}%
\begin{APACrefauthors}%
Bennett, R.E.%
\end{APACrefauthors}%
\unskip\
\newblock
\APACrefYearMonthDay{2011}{}{}.
\newblock
{\BBOQ}\APACrefatitle {Formative assessment: A critical review} {Formative assessment: A critical review}.{\BBCQ}
\newblock
\APACjournalVolNumPages{Assessment in education: principles, policy \& practice}{18}{1}{5--25,}
\newblock

\newblock

\PrintBackRefs{\CurrentBib}

\bibitem [\protect \citeauthoryear {%
Bickley%
\ \BBA {} Szilagyi%
}{%
Bickley%
\ \BBA {} Szilagyi%
}{%
{\protect \APACyear {2012}}%
}]{%
bickley2012bates}
\APACinsertmetastar {%
bickley2012bates}%
\begin{APACrefauthors}%
Bickley, L.%
\BCBT {}\ \BBA {} Szilagyi, P.G.%
\end{APACrefauthors}%
\unskip\
\newblock
\APACrefYear{2012}.
\newblock
\APACrefbtitle {Bates' guide to physical examination and history-taking} {Bates' guide to physical examination and history-taking}.
\newblock
\APACaddressPublisher{}{Lippincott Williams \& Wilkins}.
\PrintBackRefs{\CurrentBib}

\bibitem [\protect \citeauthoryear {%
Black%
\ \BBA {} Wiliam%
}{%
Black%
\ \BBA {} Wiliam%
}{%
{\protect \APACyear {1998}}%
}]{%
black1998assessment}
\APACinsertmetastar {%
black1998assessment}%
\begin{APACrefauthors}%
Black, P.%
\BCBT {}\ \BBA {} Wiliam, D.%
\end{APACrefauthors}%
\unskip\
\newblock
\APACrefYearMonthDay{1998}{}{}.
\newblock
{\BBOQ}\APACrefatitle {Assessment and classroom learning} {Assessment and classroom learning}.{\BBCQ}
\newblock
\APACjournalVolNumPages{Assessment in Education: principles, policy \& practice}{5}{1}{7--74,}
\newblock

\newblock

\PrintBackRefs{\CurrentBib}

\bibitem [\protect \citeauthoryear {%
Bond%
\ \protect \BOthers {.}}{%
Bond%
\ \protect \BOthers {.}}{%
{\protect \APACyear {2024}}%
}]{%
bond2024meta}
\APACinsertmetastar {%
bond2024meta}%
\begin{APACrefauthors}%
Bond, M.%
, Khosravi, H.%
, De~Laat, M.%
, Bergdahl, N.%
, Negrea, V.%
, Oxley, E.%
\BDBL {}Siemens, G.%
\end{APACrefauthors}%
\unskip\
\newblock
\APACrefYearMonthDay{2024}{}{}.
\newblock
{\BBOQ}\APACrefatitle {A meta systematic review of artificial intelligence in higher education: A call for increased ethics, collaboration, and rigour} {A meta systematic review of artificial intelligence in higher education: A call for increased ethics, collaboration, and rigour}.{\BBCQ}
\newblock
\APACjournalVolNumPages{International journal of educational technology in higher education}{21}{1}{4,}
\newblock

\newblock

\PrintBackRefs{\CurrentBib}

\bibitem [\protect \citeauthoryear {%
Bowen%
}{%
Bowen%
}{%
{\protect \APACyear {2006}}%
}]{%
bowen2006educational}
\APACinsertmetastar {%
bowen2006educational}%
\begin{APACrefauthors}%
Bowen, J.L.%
\end{APACrefauthors}%
\unskip\
\newblock
\APACrefYearMonthDay{2006}{}{}.
\newblock
{\BBOQ}\APACrefatitle {Educational strategies to promote clinical diagnostic reasoning} {Educational strategies to promote clinical diagnostic reasoning}.{\BBCQ}
\newblock
\APACjournalVolNumPages{New England Journal of Medicine}{355}{21}{2217--2225,}
\newblock

\newblock

\PrintBackRefs{\CurrentBib}

\bibitem [\protect \citeauthoryear {%
Br{\"u}gge%
\ \protect \BOthers {.}}{%
Br{\"u}gge%
\ \protect \BOthers {.}}{%
{\protect \APACyear {2024}}%
}]{%
brugge2024large}
\APACinsertmetastar {%
brugge2024large}%
\begin{APACrefauthors}%
Br{\"u}gge, E.%
, Ricchizzi, S.%
, Arenbeck, M.%
, Keller, M.N.%
, Schur, L.%
, Stummer, W.%
\BDBL {}Darici, D.%
\end{APACrefauthors}%
\unskip\
\newblock
\APACrefYearMonthDay{2024}{}{}.
\newblock
{\BBOQ}\APACrefatitle {Large language models improve clinical decision making of medical students through patient simulation and structured feedback: a randomized controlled trial} {Large language models improve clinical decision making of medical students through patient simulation and structured feedback: a randomized controlled trial}.{\BBCQ}
\newblock
\APACjournalVolNumPages{BMC medical education}{24}{1}{1391,}
\newblock

\newblock

\PrintBackRefs{\CurrentBib}

\bibitem [\protect \citeauthoryear {%
Charlin%
, Boshuizen%
, Custers%
\BCBL {}\ \BBA {} Feltovich%
}{%
Charlin%
\ \protect \BOthers {.}}{%
{\protect \APACyear {2007}}%
}]{%
charlin2007scripts}
\APACinsertmetastar {%
charlin2007scripts}%
\begin{APACrefauthors}%
Charlin, B.%
, Boshuizen, H.P.%
, Custers, E.J.%
\BCBL {} Feltovich, P.J.%
\end{APACrefauthors}%
\unskip\
\newblock
\APACrefYearMonthDay{2007}{}{}.
\newblock
{\BBOQ}\APACrefatitle {Scripts and clinical reasoning} {Scripts and clinical reasoning}.{\BBCQ}
\newblock
\APACjournalVolNumPages{Medical education}{41}{12}{1178--1184,}
\newblock

\newblock

\PrintBackRefs{\CurrentBib}

\bibitem [\protect \citeauthoryear {%
Csanadi%
, Eagan%
, Kollar%
, Shaffer%
\BCBL {}\ \BBA {} Fischer%
}{%
Csanadi%
\ \protect \BOthers {.}}{%
{\protect \APACyear {2018}}%
}]{%
csanadi2018coding}
\APACinsertmetastar {%
csanadi2018coding}%
\begin{APACrefauthors}%
Csanadi, A.%
, Eagan, B.%
, Kollar, I.%
, Shaffer, D.W.%
\BCBL {} Fischer, F.%
\end{APACrefauthors}%
\unskip\
\newblock
\APACrefYearMonthDay{2018}{}{}.
\newblock
{\BBOQ}\APACrefatitle {When coding-and-counting is not enough: Using epistemic network analysis (ENA) to analyze verbal data in CSCL research} {When coding-and-counting is not enough: Using epistemic network analysis (ena) to analyze verbal data in cscl research}.{\BBCQ}
\newblock
\APACjournalVolNumPages{International Journal of Computer-Supported Collaborative Learning}{13}{4}{419--438,}
\newblock

\newblock

\PrintBackRefs{\CurrentBib}

\bibitem [\protect \citeauthoryear {%
Cutrer%
\ \protect \BOthers {.}}{%
Cutrer%
\ \protect \BOthers {.}}{%
{\protect \APACyear {2017}}%
}]{%
cutrer2017fostering}
\APACinsertmetastar {%
cutrer2017fostering}%
\begin{APACrefauthors}%
Cutrer, W.B.%
, Miller, B.%
, Pusic, M.V.%
, Mejicano, G.%
, Mangrulkar, R.S.%
, Gruppen, L.D.%
\BDBL {}Moore~Jr, D.E.%
\end{APACrefauthors}%
\unskip\
\newblock
\APACrefYearMonthDay{2017}{}{}.
\newblock
{\BBOQ}\APACrefatitle {Fostering the development of master adaptive learners: a conceptual model to guide skill acquisition in medical education} {Fostering the development of master adaptive learners: a conceptual model to guide skill acquisition in medical education}.{\BBCQ}
\newblock
\APACjournalVolNumPages{Academic medicine}{92}{1}{70--75,}
\newblock

\newblock

\PrintBackRefs{\CurrentBib}

\bibitem [\protect \citeauthoryear {%
Eva%
}{%
Eva%
}{%
{\protect \APACyear {2005}}%
}]{%
eva2005every}
\APACinsertmetastar {%
eva2005every}%
\begin{APACrefauthors}%
Eva, K.W.%
\end{APACrefauthors}%
\unskip\
\newblock
\APACrefYearMonthDay{2005}{}{}.
\newblock
{\BBOQ}\APACrefatitle {What every teacher needs to know about clinical reasoning} {What every teacher needs to know about clinical reasoning}.{\BBCQ}
\newblock
\APACjournalVolNumPages{Medical education}{39}{1}{98--106,}
\newblock

\newblock

\PrintBackRefs{\CurrentBib}

\bibitem [\protect \citeauthoryear {%
F{\k{a}}ferek%
\ \protect \BOthers {.}}{%
F{\k{a}}ferek%
\ \protect \BOthers {.}}{%
{\protect \APACyear {2024}}%
}]{%
faferek2024integrating}
\APACinsertmetastar {%
faferek2024integrating}%
\begin{APACrefauthors}%
F{\k{a}}ferek, J.%
, Cariou, P\BHBI L.%
, Hege, I.%
, Mayer, A.%
, Morin, L.%
, Rodriguez-Molina, D.%
\BDBL {}Kononowicz, A.A.%
\end{APACrefauthors}%
\unskip\
\newblock
\APACrefYearMonthDay{2024}{}{}.
\newblock
{\BBOQ}\APACrefatitle {Integrating virtual patients into undergraduate health professions curricula: a framework synthesis of stakeholders’ opinions based on a systematic literature review} {Integrating virtual patients into undergraduate health professions curricula: a framework synthesis of stakeholders’ opinions based on a systematic literature review}.{\BBCQ}
\newblock
\APACjournalVolNumPages{BMC Medical Education}{24}{1}{727,}
\newblock

\newblock

\PrintBackRefs{\CurrentBib}

\bibitem [\protect \citeauthoryear {%
F{\"u}rstenberg%
\ \protect \BOthers {.}}{%
F{\"u}rstenberg%
\ \protect \BOthers {.}}{%
{\protect \APACyear {2020}}%
}]{%
furstenberg2020assessing}
\APACinsertmetastar {%
furstenberg2020assessing}%
\begin{APACrefauthors}%
F{\"u}rstenberg, S.%
, Helm, T.%
, Prediger, S.%
, Kadmon, M.%
, Berberat, P.O.%
\BCBL {} Harendza, S.%
\end{APACrefauthors}%
\unskip\
\newblock
\APACrefYearMonthDay{2020}{}{}.
\newblock
{\BBOQ}\APACrefatitle {Assessing clinical reasoning in undergraduate medical students during history taking with an empirically derived scale for clinical reasoning indicators} {Assessing clinical reasoning in undergraduate medical students during history taking with an empirically derived scale for clinical reasoning indicators}.{\BBCQ}
\newblock
\APACjournalVolNumPages{BMC Medical Education}{20}{1}{368,}
\newblock

\newblock

\PrintBackRefs{\CurrentBib}

\bibitem [\protect \citeauthoryear {%
Goldowsky%
\ \BBA {} Rencic%
}{%
Goldowsky%
\ \BBA {} Rencic%
}{%
{\protect \APACyear {2023}}%
}]{%
goldowsky2023self}
\APACinsertmetastar {%
goldowsky2023self}%
\begin{APACrefauthors}%
Goldowsky, A.%
\BCBT {}\ \BBA {} Rencic, J.%
\end{APACrefauthors}%
\unskip\
\newblock
\APACrefYearMonthDay{2023}{}{}.
\newblock
{\BBOQ}\APACrefatitle {Self-regulated learning and the future of diagnostic reasoning education} {Self-regulated learning and the future of diagnostic reasoning education}.{\BBCQ}
\newblock
\APACjournalVolNumPages{Diagnosis}{10}{1}{24--30,}
\newblock

\newblock

\PrintBackRefs{\CurrentBib}

\bibitem [\protect \citeauthoryear {%
Hamilton%
, Molzahn%
\BCBL {}\ \BBA {} McLemore%
}{%
Hamilton%
\ \protect \BOthers {.}}{%
{\protect \APACyear {2024}}%
}]{%
hamilton2024evolution}
\APACinsertmetastar {%
hamilton2024evolution}%
\begin{APACrefauthors}%
Hamilton, A.%
, Molzahn, A.%
\BCBL {} McLemore, K.%
\end{APACrefauthors}%
\unskip\
\newblock
\APACrefYearMonthDay{2024}{}{}.
\newblock
{\BBOQ}\APACrefatitle {The evolution from standardized to virtual patients in medical education} {The evolution from standardized to virtual patients in medical education}.{\BBCQ}
\newblock
\APACjournalVolNumPages{Cureus}{16}{10}{,}
\newblock

\newblock

\PrintBackRefs{\CurrentBib}

\bibitem [\protect \citeauthoryear {%
Haring%
, Cools%
, van Gurp%
, van~der Meer%
\BCBL {}\ \BBA {} Postma%
}{%
Haring%
\ \protect \BOthers {.}}{%
{\protect \APACyear {2017}}%
}]{%
haring2017observable}
\APACinsertmetastar {%
haring2017observable}%
\begin{APACrefauthors}%
Haring, C.M.%
, Cools, B.M.%
, van Gurp, P.J.%
, van~der Meer, J.W.%
\BCBL {} Postma, C.T.%
\end{APACrefauthors}%
\unskip\
\newblock
\APACrefYearMonthDay{2017}{}{}.
\newblock
{\BBOQ}\APACrefatitle {Observable phenomena that reveal medical students' clinical reasoning ability during expert assessment of their history taking: a qualitative study} {Observable phenomena that reveal medical students' clinical reasoning ability during expert assessment of their history taking: a qualitative study}.{\BBCQ}
\newblock
\APACjournalVolNumPages{BMC Medical education}{17}{1}{147,}
\newblock

\newblock

\PrintBackRefs{\CurrentBib}

\bibitem [\protect \citeauthoryear {%
Haring%
\ \protect \BOthers {.}}{%
Haring%
\ \protect \BOthers {.}}{%
{\protect \APACyear {2020}}%
}]{%
haring2020validity}
\APACinsertmetastar {%
haring2020validity}%
\begin{APACrefauthors}%
Haring, C.M.%
, Klaarwater, C.C.%
, Bouwmans, G.A.%
, Cools, B.M.%
, van Gurp, P.J.%
, van~der Meer, J.W.%
\BCBL {} Postma, C.T.%
\end{APACrefauthors}%
\unskip\
\newblock
\APACrefYearMonthDay{2020}{}{}.
\newblock
{\BBOQ}\APACrefatitle {Validity, reliability and feasibility of a new observation rating tool and a post encounter rating tool for the assessment of clinical reasoning skills of medical students during their internal medicine clerkship: a pilot study} {Validity, reliability and feasibility of a new observation rating tool and a post encounter rating tool for the assessment of clinical reasoning skills of medical students during their internal medicine clerkship: a pilot study}.{\BBCQ}
\newblock
\APACjournalVolNumPages{BMC Medical Education}{20}{1}{198,}
\newblock

\newblock

\PrintBackRefs{\CurrentBib}

\bibitem [\protect \citeauthoryear {%
Hasnain%
, Bordage%
, Connell%
\BCBL {}\ \BBA {} Sinacore%
}{%
Hasnain%
\ \protect \BOthers {.}}{%
{\protect \APACyear {2001}}%
}]{%
hasnain2001historytaking}
\APACinsertmetastar {%
hasnain2001historytaking}%
\begin{APACrefauthors}%
Hasnain, M.%
, Bordage, G.%
, Connell, K.J.%
\BCBL {} Sinacore, J.M.%
\end{APACrefauthors}%
\unskip\
\newblock
\APACrefYearMonthDay{2001}{}{}.
\newblock
{\BBOQ}\APACrefatitle {History-taking behaviors associated with diagnostic competence of clerks: an exploratory study} {History-taking behaviors associated with diagnostic competence of clerks: an exploratory study}.{\BBCQ}
\newblock
\APACjournalVolNumPages{Academic Medicine}{76}{10}{S14--S17,}
\newblock

\newblock

\PrintBackRefs{\CurrentBib}

\bibitem [\protect \citeauthoryear {%
Hege%
, Kononowicz%
, Berman%
, Lenzer%
\BCBL {}\ \BBA {} Kiesewetter%
}{%
Hege%
\ \protect \BOthers {.}}{%
{\protect \APACyear {2018}}%
}]{%
hege2018advancing}
\APACinsertmetastar {%
hege2018advancing}%
\begin{APACrefauthors}%
Hege, I.%
, Kononowicz, A.A.%
, Berman, N.B.%
, Lenzer, B.%
\BCBL {} Kiesewetter, J.%
\end{APACrefauthors}%
\unskip\
\newblock
\APACrefYearMonthDay{2018}{}{}.
\newblock
{\BBOQ}\APACrefatitle {Advancing clinical reasoning in virtual patients--development and application of a conceptual framework} {Advancing clinical reasoning in virtual patients--development and application of a conceptual framework}.{\BBCQ}
\newblock
\APACjournalVolNumPages{GMS journal for medical education}{35}{1}{Doc12,}
\newblock

\newblock

\PrintBackRefs{\CurrentBib}

\bibitem [\protect \citeauthoryear {%
Holderried%
, Stegemann-Philipps%
, Herrmann-Werner%
\BCBL {}\ \protect \BOthers {.}}{%
Holderried%
, Stegemann-Philipps%
, Herrmann-Werner%
\BCBL {}\ \protect \BOthers {.}}{%
{\protect \APACyear {2024}}%
}]{%
holderried2024feedback}
\APACinsertmetastar {%
holderried2024feedback}%
\begin{APACrefauthors}%
Holderried, F.%
, Stegemann-Philipps, C.%
, Herrmann-Werner, A.%
, Festl-Wietek, T.%
, Holderried, M.%
, Eickhoff, C.%
\BDBL {}others%
\end{APACrefauthors}%
\unskip\
\newblock
\APACrefYearMonthDay{2024}{}{}.
\newblock
{\BBOQ}\APACrefatitle {A language model--powered simulated patient with automated feedback for history taking: Prospective study} {A language model--powered simulated patient with automated feedback for history taking: Prospective study}.{\BBCQ}
\newblock
\APACjournalVolNumPages{JMIR Medical Education}{10}{1}{e59213,}
\newblock

\newblock

\PrintBackRefs{\CurrentBib}

\bibitem [\protect \citeauthoryear {%
Holderried%
, Stegemann-Philipps%
, Herschbach%
\BCBL {}\ \protect \BOthers {.}}{%
Holderried%
, Stegemann-Philipps%
, Herschbach%
\BCBL {}\ \protect \BOthers {.}}{%
{\protect \APACyear {2024}}%
}]{%
holderried2024generative}
\APACinsertmetastar {%
holderried2024generative}%
\begin{APACrefauthors}%
Holderried, F.%
, Stegemann-Philipps, C.%
, Herschbach, L.%
, Moldt, J\BHBI A.%
, Nevins, A.%
, Griewatz, J.%
\BDBL {}Mahling, M.%
\end{APACrefauthors}%
\unskip\
\newblock
\APACrefYearMonthDay{2024}{}{}.
\newblock
{\BBOQ}\APACrefatitle {A generative pretrained transformer (GPT)--powered chatbot as a simulated patient to practice history taking: prospective, mixed methods study} {A generative pretrained transformer (gpt)--powered chatbot as a simulated patient to practice history taking: prospective, mixed methods study}.{\BBCQ}
\newblock
\APACjournalVolNumPages{JMIR medical education}{10}{1}{e53961,}
\newblock

\newblock

\PrintBackRefs{\CurrentBib}

\bibitem [\protect \citeauthoryear {%
J{\"a}rvel{\"a}%
, Nguyen%
\BCBL {}\ \BBA {} Hadwin%
}{%
J{\"a}rvel{\"a}%
\ \protect \BOthers {.}}{%
{\protect \APACyear {2023}}%
}]{%
jarvela2023human}
\APACinsertmetastar {%
jarvela2023human}%
\begin{APACrefauthors}%
J{\"a}rvel{\"a}, S.%
, Nguyen, A.%
\BCBL {} Hadwin, A.%
\end{APACrefauthors}%
\unskip\
\newblock
\APACrefYearMonthDay{2023}{}{}.
\newblock
{\BBOQ}\APACrefatitle {Human and artificial intelligence collaboration for socially shared regulation in learning} {Human and artificial intelligence collaboration for socially shared regulation in learning}.{\BBCQ}
\newblock
\APACjournalVolNumPages{British Journal of Educational Technology}{54}{5}{1057--1076,}
\newblock
\begin{APACrefDOI} \doi{10.1111/bjet.13325} \end{APACrefDOI}
\newblock

\newblock

\PrintBackRefs{\CurrentBib}

\bibitem [\protect \citeauthoryear {%
Jay%
\ \protect \BOthers {.}}{%
Jay%
\ \protect \BOthers {.}}{%
{\protect \APACyear {2025}}%
}]{%
jay2025use}
\APACinsertmetastar {%
jay2025use}%
\begin{APACrefauthors}%
Jay, R.%
, Sandars, J.%
, Patel, R.%
, Leonardi-Bee, J.%
, Ackbarally, Y.%
, Bandyopadhyay, S.%
\BDBL {}Wilson, E.%
\end{APACrefauthors}%
\unskip\
\newblock
\APACrefYearMonthDay{2025}{}{}.
\newblock
{\BBOQ}\APACrefatitle {The use of virtual patients to provide feedback on clinical reasoning: a systematic review} {The use of virtual patients to provide feedback on clinical reasoning: a systematic review}.{\BBCQ}
\newblock
\APACjournalVolNumPages{Academic Medicine}{100}{2}{229--238,}
\newblock

\newblock

\PrintBackRefs{\CurrentBib}

\bibitem [\protect \citeauthoryear {%
Keifenheim%
\ \protect \BOthers {.}}{%
Keifenheim%
\ \protect \BOthers {.}}{%
{\protect \APACyear {2015}}%
}]{%
keifenheim2015teaching}
\APACinsertmetastar {%
keifenheim2015teaching}%
\begin{APACrefauthors}%
Keifenheim, K.E.%
, Teufel, M.%
, Ip, J.%
, Speiser, N.%
, Leehr, E.J.%
, Zipfel, S.%
\BCBL {} Herrmann-Werner, A.%
\end{APACrefauthors}%
\unskip\
\newblock
\APACrefYearMonthDay{2015}{}{}.
\newblock
{\BBOQ}\APACrefatitle {Teaching history taking to medical students: a systematic review} {Teaching history taking to medical students: a systematic review}.{\BBCQ}
\newblock
\APACjournalVolNumPages{BMC medical education}{15}{1}{159,}
\newblock

\newblock

\PrintBackRefs{\CurrentBib}

\bibitem [\protect \citeauthoryear {%
Kononowicz%
\ \protect \BOthers {.}}{%
Kononowicz%
\ \protect \BOthers {.}}{%
{\protect \APACyear {2019}}%
}]{%
kononowicz2019virtual}
\APACinsertmetastar {%
kononowicz2019virtual}%
\begin{APACrefauthors}%
Kononowicz, A.A.%
, Woodham, L.A.%
, Edelbring, S.%
, Stathakarou, N.%
, Davies, D.%
, Saxena, N.%
\BDBL {}Zary, N.%
\end{APACrefauthors}%
\unskip\
\newblock
\APACrefYearMonthDay{2019}{}{}.
\newblock
{\BBOQ}\APACrefatitle {Virtual patient simulations in health professions education: systematic review and meta-analysis by the digital health education collaboration} {Virtual patient simulations in health professions education: systematic review and meta-analysis by the digital health education collaboration}.{\BBCQ}
\newblock
\APACjournalVolNumPages{Journal of medical Internet research}{21}{7}{e14676,}
\newblock

\newblock

\PrintBackRefs{\CurrentBib}

\bibitem [\protect \citeauthoryear {%
Kurtz%
, Silverman%
, Benson%
\BCBL {}\ \BBA {} Draper%
}{%
Kurtz%
\ \protect \BOthers {.}}{%
{\protect \APACyear {2003}}%
}]{%
kurtz2003marrying}
\APACinsertmetastar {%
kurtz2003marrying}%
\begin{APACrefauthors}%
Kurtz, S.%
, Silverman, J.%
, Benson, J.%
\BCBL {} Draper, J.%
\end{APACrefauthors}%
\unskip\
\newblock
\APACrefYearMonthDay{2003}{}{}.
\newblock
{\BBOQ}\APACrefatitle {Marrying content and process in clinical method teaching: Enhancing the Calgary-Cambridge guides} {Marrying content and process in clinical method teaching: Enhancing the calgary-cambridge guides}.{\BBCQ}
\newblock
\APACjournalVolNumPages{Academic Medicine}{78}{8}{802--809,}
\newblock
\begin{APACrefDOI} \doi{10.1097/00001888-200308000-00011} \end{APACrefDOI}
\newblock

\newblock

\PrintBackRefs{\CurrentBib}

\bibitem [\protect \citeauthoryear {%
D.~Li%
\ \BBA {} {Lebai Lutfi}%
}{%
D.~Li%
\ \BBA {} {Lebai Lutfi}%
}{%
{\protect \APACyear {2026}}%
}]{%
li2026large}
\APACinsertmetastar {%
li2026large}%
\begin{APACrefauthors}%
Li, D.%
\BCBT {}\ \BBA {} {Lebai Lutfi}, S.%
\end{APACrefauthors}%
\unskip\
\newblock
\APACrefYearMonthDay{2026}{}{}.
\newblock
{\BBOQ}\APACrefatitle {Large Language Model--Based Virtual Patient Systems for History-Taking in Medical Education: Comprehensive Systematic Review} {Large language model--based virtual patient systems for history-taking in medical education: Comprehensive systematic review}.{\BBCQ}
\newblock
\APACjournalVolNumPages{JMIR Medical Informatics}{14}{}{e79039,}
\newblock
\begin{APACrefDOI} \doi{10.2196/79039} \end{APACrefDOI}
\newblock
\begin{APACrefURL} {https://medinform.jmir.org/2026/1/e79039/} \end{APACrefURL}
\newblock

\newblock

\PrintBackRefs{\CurrentBib}

\bibitem [\protect \citeauthoryear {%
T.~Li%
\ \protect \BOthers {.}}{%
T.~Li%
\ \protect \BOthers {.}}{%
{\protect \APACyear {2023}}%
}]{%
li2023analytics}
\APACinsertmetastar {%
li2023analytics}%
\begin{APACrefauthors}%
Li, T.%
, Fan, Y.%
, Tan, Y.%
, Wang, Y.%
, Singh, S.%
, Li, X.%
\BDBL {}others%
\end{APACrefauthors}%
\unskip\
\newblock
\APACrefYearMonthDay{2023}{}{}.
\newblock
{\BBOQ}\APACrefatitle {Analytics of self-regulated learning scaffolding: effects on learning processes} {Analytics of self-regulated learning scaffolding: effects on learning processes}.{\BBCQ}
\newblock
\APACjournalVolNumPages{Frontiers in Psychology}{14}{}{,}
\newblock

\newblock

\PrintBackRefs{\CurrentBib}

\bibitem [\protect \citeauthoryear {%
Mahbubani%
}{%
Mahbubani%
}{%
{\protect \APACyear {2023}}%
}]{%
mahbubani2023history}
\APACinsertmetastar {%
mahbubani2023history}%
\begin{APACrefauthors}%
Mahbubani, K.%
\end{APACrefauthors}%
\unskip\
\newblock
\APACrefYear{2023}.
\newblock
\APACrefbtitle {History Taking in Clinical Practice} {History taking in clinical practice}.
\newblock
\APACaddressPublisher{}{Springer}.
\PrintBackRefs{\CurrentBib}

\bibitem [\protect \citeauthoryear {%
Maicher%
\ \protect \BOthers {.}}{%
Maicher%
\ \protect \BOthers {.}}{%
{\protect \APACyear {2023}}%
}]{%
maicher2023artificial}
\APACinsertmetastar {%
maicher2023artificial}%
\begin{APACrefauthors}%
Maicher, K.R.%
, Stiff, A.%
, Scholl, M.%
, White, M.%
, Fosler-Lussier, E.%
, Schuler, W.%
\BDBL {}others%
\end{APACrefauthors}%
\unskip\
\newblock
\APACrefYearMonthDay{2023}{}{}.
\newblock
{\BBOQ}\APACrefatitle {Artificial intelligence in virtual standardized patients: combining natural language understanding and rule based dialogue management to improve conversational fidelity} {Artificial intelligence in virtual standardized patients: combining natural language understanding and rule based dialogue management to improve conversational fidelity}.{\BBCQ}
\newblock
\APACjournalVolNumPages{Medical teacher}{45}{3}{279--285,}
\newblock

\newblock

\PrintBackRefs{\CurrentBib}

\bibitem [\protect \citeauthoryear {%
Makoul%
}{%
Makoul%
}{%
{\protect \APACyear {2001}}%
}]{%
makoul2001essential}
\APACinsertmetastar {%
makoul2001essential}%
\begin{APACrefauthors}%
Makoul, G.%
\end{APACrefauthors}%
\unskip\
\newblock
\APACrefYearMonthDay{2001}{}{}.
\newblock
{\BBOQ}\APACrefatitle {Essential elements of communication in medical encounters: The Kalamazoo consensus statement} {Essential elements of communication in medical encounters: The kalamazoo consensus statement}.{\BBCQ}
\newblock
\APACjournalVolNumPages{Academic Medicine}{76}{4}{390--393,}
\newblock
\begin{APACrefDOI} \doi{10.1097/00001888-200104000-00021} \end{APACrefDOI}
\newblock

\newblock

\PrintBackRefs{\CurrentBib}

\bibitem [\protect \citeauthoryear {%
Malau-Aduli%
, Jones%
, Saad%
\BCBL {}\ \BBA {} Richmond%
}{%
Malau-Aduli%
\ \protect \BOthers {.}}{%
{\protect \APACyear {2022}}%
}]{%
malauaduli2022osce}
\APACinsertmetastar {%
malauaduli2022osce}%
\begin{APACrefauthors}%
Malau-Aduli, B.S.%
, Jones, K.%
, Saad, S.%
\BCBL {} Richmond, C.%
\end{APACrefauthors}%
\unskip\
\newblock
\APACrefYearMonthDay{2022}{}{}.
\newblock
{\BBOQ}\APACrefatitle {Has the OSCE met its final demise? Rebalancing clinical assessment approaches in the peri-pandemic world} {Has the osce met its final demise? rebalancing clinical assessment approaches in the peri-pandemic world}.{\BBCQ}
\newblock
\APACjournalVolNumPages{Frontiers in Medicine}{9}{}{825502,}
\newblock
\begin{APACrefDOI} \doi{10.3389/fmed.2022.825502} \end{APACrefDOI}
\newblock

\newblock

\PrintBackRefs{\CurrentBib}

\bibitem [\protect \citeauthoryear {%
Matcha%
, Ga{\v{s}}evi{\'c}%
, Uzir%
, Jovanovi{\'c}%
\BCBL {}\ \BBA {} Pardo%
}{%
Matcha%
\ \protect \BOthers {.}}{%
{\protect \APACyear {2019}}%
}]{%
matcha2019analytics}
\APACinsertmetastar {%
matcha2019analytics}%
\begin{APACrefauthors}%
Matcha, W.%
, Ga{\v{s}}evi{\'c}, D.%
, Uzir, N.A.%
, Jovanovi{\'c}, J.%
\BCBL {} Pardo, A.%
\end{APACrefauthors}%
\unskip\
\newblock
\APACrefYearMonthDay{2019}{}{}.
\newblock
{\BBOQ}\APACrefatitle {Analytics of learning strategies: Associations with academic performance and feedback} {Analytics of learning strategies: Associations with academic performance and feedback}.{\BBCQ}
\newblock
 \APACrefbtitle {Proceedings of the 9th International Conference on Learning Analytics \& Knowledge} {Proceedings of the 9th international conference on learning analytics \& knowledge}\ (\BPGS\ 461--470).
\PrintBackRefs{\CurrentBib}

\bibitem [\protect \citeauthoryear {%
Milota%
, van Thiel%
\BCBL {}\ \BBA {} van Delden%
}{%
Milota%
\ \protect \BOthers {.}}{%
{\protect \APACyear {2019}}%
}]{%
milota2019narrative}
\APACinsertmetastar {%
milota2019narrative}%
\begin{APACrefauthors}%
Milota, M.M.%
, van Thiel, G.J.M.W.%
\BCBL {} van Delden, J.J.M.%
\end{APACrefauthors}%
\unskip\
\newblock
\APACrefYearMonthDay{2019}{}{}.
\newblock
{\BBOQ}\APACrefatitle {Narrative medicine as a medical education tool: A systematic review} {Narrative medicine as a medical education tool: A systematic review}.{\BBCQ}
\newblock
\APACjournalVolNumPages{Medical Teacher}{41}{7}{802--810,}
\newblock
\begin{APACrefDOI} \doi{10.1080/0142159X.2019.1584274} \end{APACrefDOI}
\newblock

\newblock

\PrintBackRefs{\CurrentBib}

\bibitem [\protect \citeauthoryear {%
Misra%
\ \BBA {} Suresh%
}{%
Misra%
\ \BBA {} Suresh%
}{%
{\protect \APACyear {2024}}%
}]{%
misra2024osce}
\APACinsertmetastar {%
misra2024osce}%
\begin{APACrefauthors}%
Misra, S.M.%
\BCBT {}\ \BBA {} Suresh, S.%
\end{APACrefauthors}%
\unskip\
\newblock
\APACrefYearMonthDay{2024}{}{}.
\newblock
{\BBOQ}\APACrefatitle {Artificial intelligence and objective structured clinical examinations: using ChatGPT to revolutionize clinical skills assessment in medical education} {Artificial intelligence and objective structured clinical examinations: using chatgpt to revolutionize clinical skills assessment in medical education}.{\BBCQ}
\newblock
\APACjournalVolNumPages{Journal of Medical Education and Curricular Development}{11}{}{23821205241263475,}
\newblock
\begin{APACrefDOI} \doi{10.1177/23821205241263475} \end{APACrefDOI}
\newblock

\newblock

\PrintBackRefs{\CurrentBib}

\bibitem [\protect \citeauthoryear {%
Molenaar%
\ \protect \BOthers {.}}{%
Molenaar%
\ \protect \BOthers {.}}{%
{\protect \APACyear {2023}}%
}]{%
molenaar2023measuring}
\APACinsertmetastar {%
molenaar2023measuring}%
\begin{APACrefauthors}%
Molenaar, I.%
, de Mooij, S.%
, Azevedo, R.%
, Bannert, M.%
, J{\"a}rvel{\"a}, S.%
\BCBL {} Ga{\v{s}}evi{\'c}, D.%
\end{APACrefauthors}%
\unskip\
\newblock
\APACrefYearMonthDay{2023}{}{}.
\newblock
{\BBOQ}\APACrefatitle {Measuring self-regulated learning and the role of AI: Five years of research using multimodal multichannel data} {Measuring self-regulated learning and the role of ai: Five years of research using multimodal multichannel data}.{\BBCQ}
\newblock
\APACjournalVolNumPages{Computers in Human Behavior}{139}{}{107540,}
\newblock

\newblock

\PrintBackRefs{\CurrentBib}

\bibitem [\protect \citeauthoryear {%
Nendaz%
\ \protect \BOthers {.}}{%
Nendaz%
\ \protect \BOthers {.}}{%
{\protect \APACyear {2006}}%
}]{%
nendaz2006beyond}
\APACinsertmetastar {%
nendaz2006beyond}%
\begin{APACrefauthors}%
Nendaz, M.R.%
, Gut, A.M.%
, Perrier, A.%
, Louis-Simonet, M.%
, Blondon-Choa, K.%
, Herrmann, F.R.%
\BDBL {}Vu, N.V.%
\end{APACrefauthors}%
\unskip\
\newblock
\APACrefYearMonthDay{2006}{}{}.
\newblock
{\BBOQ}\APACrefatitle {Brief report: beyond clinical experience: features of data collection and interpretation that contribute to diagnostic accuracy} {Brief report: beyond clinical experience: features of data collection and interpretation that contribute to diagnostic accuracy}.{\BBCQ}
\newblock
\APACjournalVolNumPages{Journal of general internal medicine}{21}{12}{1302--1305,}
\newblock

\newblock

\PrintBackRefs{\CurrentBib}

\bibitem [\protect \citeauthoryear {%
Ng%
\ \protect \BOthers {.}}{%
Ng%
\ \protect \BOthers {.}}{%
{\protect \APACyear {2025}}%
}]{%
ng2025clinical}
\APACinsertmetastar {%
ng2025clinical}%
\begin{APACrefauthors}%
Ng, I.K.S.%
, Goh, W.G.W.%
, Teo, D.B.%
, Chong, K.M.%
, Tan, L.F.%
\BCBL {} Teoh, C.M.%
\end{APACrefauthors}%
\unskip\
\newblock
\APACrefYearMonthDay{2025}{}{}.
\newblock
{\BBOQ}\APACrefatitle {Clinical reasoning in real-world practice: a primer for medical trainees and practitioners} {Clinical reasoning in real-world practice: a primer for medical trainees and practitioners}.{\BBCQ}
\newblock
\APACjournalVolNumPages{Postgraduate Medical Journal}{101}{1191}{68--75,}
\newblock
\begin{APACrefDOI} \doi{10.1093/postmj/qgae079} \end{APACrefDOI}
\newblock

\newblock

\PrintBackRefs{\CurrentBib}

\bibitem [\protect \citeauthoryear {%
Plackett%
\ \protect \BOthers {.}}{%
Plackett%
\ \protect \BOthers {.}}{%
{\protect \APACyear {2022}}%
}]{%
plackett2022effectiveness}
\APACinsertmetastar {%
plackett2022effectiveness}%
\begin{APACrefauthors}%
Plackett, R.%
, Kassianos, A.P.%
, Mylan, S.%
, Kambouri, M.%
, Raine, R.%
\BCBL {} Sheringham, J.%
\end{APACrefauthors}%
\unskip\
\newblock
\APACrefYearMonthDay{2022}{}{}.
\newblock
{\BBOQ}\APACrefatitle {The effectiveness of using virtual patient educational tools to improve medical students' clinical reasoning skills: a systematic review} {The effectiveness of using virtual patient educational tools to improve medical students' clinical reasoning skills: a systematic review}.{\BBCQ}
\newblock
\APACjournalVolNumPages{BMC Medical Education}{22}{1}{365,}
\newblock
\begin{APACrefDOI} \doi{10.1186/s12909-022-03410-x} \end{APACrefDOI}
\newblock

\newblock

\PrintBackRefs{\CurrentBib}

\bibitem [\protect \citeauthoryear {%
Rakovi{\'c}%
\ \protect \BOthers {.}}{%
Rakovi{\'c}%
\ \protect \BOthers {.}}{%
{\protect \APACyear {2023}}%
}]{%
rakovic2023harnessing}
\APACinsertmetastar {%
rakovic2023harnessing}%
\begin{APACrefauthors}%
Rakovi{\'c}, M.%
, Iqbal, S.%
, Li, T.%
, Fan, Y.%
, Singh, S.%
, Surendrannair, S.%
\BDBL {}others%
\end{APACrefauthors}%
\unskip\
\newblock
\APACrefYearMonthDay{2023}{}{}.
\newblock
{\BBOQ}\APACrefatitle {Harnessing the potential of trace data and linguistic analysis to predict learner performance in a multi-text writing task} {Harnessing the potential of trace data and linguistic analysis to predict learner performance in a multi-text writing task}.{\BBCQ}
\newblock
\APACjournalVolNumPages{Journal of Computer Assisted Learning}{39}{3}{,}
\newblock

\newblock

\PrintBackRefs{\CurrentBib}

\bibitem [\protect \citeauthoryear {%
Regehr%
, MacRae%
, Reznick%
\BCBL {}\ \BBA {} Szalay%
}{%
Regehr%
\ \protect \BOthers {.}}{%
{\protect \APACyear {1998}}%
}]{%
regehr1998comparing}
\APACinsertmetastar {%
regehr1998comparing}%
\begin{APACrefauthors}%
Regehr, G.%
, MacRae, H.%
, Reznick, R.K.%
\BCBL {} Szalay, D.%
\end{APACrefauthors}%
\unskip\
\newblock
\APACrefYearMonthDay{1998}{}{}.
\newblock
{\BBOQ}\APACrefatitle {Comparing the psychometric properties of checklists and global rating scales for assessing performance on an OSCE-format examination} {Comparing the psychometric properties of checklists and global rating scales for assessing performance on an osce-format examination}.{\BBCQ}
\newblock
\APACjournalVolNumPages{Academic Medicine}{73}{9}{993--7,}
\newblock

\newblock

\PrintBackRefs{\CurrentBib}

\bibitem [\protect \citeauthoryear {%
Reimann%
}{%
Reimann%
}{%
{\protect \APACyear {2007}}%
}]{%
reimann2007time}
\APACinsertmetastar {%
reimann2007time}%
\begin{APACrefauthors}%
Reimann, P.%
\end{APACrefauthors}%
\unskip\
\newblock
\APACrefYearMonthDay{2007}{}{}.
\newblock
{\BBOQ}\APACrefatitle {Time is precious: Why process analysis is essential for CSCL (and can also help to bridge between experimental and descriptive methods)} {Time is precious: Why process analysis is essential for cscl (and can also help to bridge between experimental and descriptive methods)}.{\BBCQ}
\newblock

\newblock

\newblock

\PrintBackRefs{\CurrentBib}

\bibitem [\protect \citeauthoryear {%
Roter%
\ \BBA {} Larson%
}{%
Roter%
\ \BBA {} Larson%
}{%
{\protect \APACyear {2002}}%
}]{%
roter2002roter}
\APACinsertmetastar {%
roter2002roter}%
\begin{APACrefauthors}%
Roter, D.%
\BCBT {}\ \BBA {} Larson, S.%
\end{APACrefauthors}%
\unskip\
\newblock
\APACrefYearMonthDay{2002}{}{}.
\newblock
{\BBOQ}\APACrefatitle {The Roter interaction analysis system (RIAS): utility and flexibility for analysis of medical interactions} {The roter interaction analysis system (rias): utility and flexibility for analysis of medical interactions}.{\BBCQ}
\newblock
\APACjournalVolNumPages{Patient education and counseling}{46}{4}{243--251,}
\newblock

\newblock

\PrintBackRefs{\CurrentBib}

\bibitem [\protect \citeauthoryear {%
Saint%
, Fan%
, Ga{\v{s}}evi{\'c}%
\BCBL {}\ \BBA {} Pardo%
}{%
Saint%
\ \protect \BOthers {.}}{%
{\protect \APACyear {2022}}%
}]{%
saint2022temporally}
\APACinsertmetastar {%
saint2022temporally}%
\begin{APACrefauthors}%
Saint, J.%
, Fan, Y.%
, Ga{\v{s}}evi{\'c}, D.%
\BCBL {} Pardo, A.%
\end{APACrefauthors}%
\unskip\
\newblock
\APACrefYearMonthDay{2022}{}{}.
\newblock
{\BBOQ}\APACrefatitle {Temporally-focused analytics of self-regulated learning: A systematic review of literature} {Temporally-focused analytics of self-regulated learning: A systematic review of literature}.{\BBCQ}
\newblock
\APACjournalVolNumPages{Computers and Education: Artificial Intelligence}{3}{}{100060,}
\newblock
\begin{APACrefDOI} \doi{10.1016/j.caeai.2022.100060} \end{APACrefDOI}
\newblock

\newblock

\PrintBackRefs{\CurrentBib}

\bibitem [\protect \citeauthoryear {%
Saint%
, Ga{\v{s}}evi{\'c}%
, Matcha%
, Uzir%
\BCBL {}\ \BBA {} Pardo%
}{%
Saint%
\ \protect \BOthers {.}}{%
{\protect \APACyear {2020}}%
}]{%
saint2020combining}
\APACinsertmetastar {%
saint2020combining}%
\begin{APACrefauthors}%
Saint, J.%
, Ga{\v{s}}evi{\'c}, D.%
, Matcha, W.%
, Uzir, N.A.%
\BCBL {} Pardo, A.%
\end{APACrefauthors}%
\unskip\
\newblock
\APACrefYearMonthDay{2020}{}{}.
\newblock
{\BBOQ}\APACrefatitle {Combining analytic methods to unlock sequential and temporal patterns of self-regulated learning} {Combining analytic methods to unlock sequential and temporal patterns of self-regulated learning}.{\BBCQ}
\newblock
 \APACrefbtitle {{Proceedings of the 10th International Conference on Learning Analytics and Knowledge \emph{(LAK 2020), 23--27 March 2020, Frankfurt, Germany}}} {{Proceedings of the 10th International Conference on Learning Analytics and Knowledge \emph{(LAK 2020), 23--27 March 2020, Frankfurt, Germany}}}\ (\BPGS\ 402--411).
\newblock
\APACaddressPublisher{}{ACM}.
\PrintBackRefs{\CurrentBib}

\bibitem [\protect \citeauthoryear {%
Saqr%
\ \protect \BOthers {.}}{%
Saqr%
\ \protect \BOthers {.}}{%
{\protect \APACyear {2024}}%
}]{%
saqr2024sequence}
\APACinsertmetastar {%
saqr2024sequence}%
\begin{APACrefauthors}%
Saqr, M.%
, L{\'o}pez-Pernas, S.%
, Helske, S.%
, Durand, M.%
, Murphy, K.%
, Studer, M.%
\BCBL {} Ritschard, G.%
\end{APACrefauthors}%
\unskip\
\newblock
\APACrefYearMonthDay{2024}{}{}.
\newblock
{\BBOQ}\APACrefatitle {Sequence analysis in education: principles, technique, and tutorial with R} {Sequence analysis in education: principles, technique, and tutorial with r}.{\BBCQ}
\newblock
 \APACrefbtitle {Learning analytics methods and tutorials: A practical guide using R} {Learning analytics methods and tutorials: A practical guide using r}\ (\BPGS\ 321--354).
\newblock
\APACaddressPublisher{}{Springer Nature Switzerland Cham}.
\PrintBackRefs{\CurrentBib}

\bibitem [\protect \citeauthoryear {%
Schmidt%
\ \BBA {} Rikers%
}{%
Schmidt%
\ \BBA {} Rikers%
}{%
{\protect \APACyear {2007}}%
}]{%
schmidt2007expertise}
\APACinsertmetastar {%
schmidt2007expertise}%
\begin{APACrefauthors}%
Schmidt, H.G.%
\BCBT {}\ \BBA {} Rikers, R.M.%
\end{APACrefauthors}%
\unskip\
\newblock
\APACrefYearMonthDay{2007}{}{}.
\newblock
{\BBOQ}\APACrefatitle {How expertise develops in medicine: knowledge encapsulation and illness script formation} {How expertise develops in medicine: knowledge encapsulation and illness script formation}.{\BBCQ}
\newblock
\APACjournalVolNumPages{Medical education}{41}{12}{1133--1139,}
\newblock

\newblock

\PrintBackRefs{\CurrentBib}

\bibitem [\protect \citeauthoryear {%
Shaffer%
, Collier%
\BCBL {}\ \BBA {} Ruis%
}{%
Shaffer%
\ \protect \BOthers {.}}{%
{\protect \APACyear {2016}}%
}]{%
shaffer2016tutorial}
\APACinsertmetastar {%
shaffer2016tutorial}%
\begin{APACrefauthors}%
Shaffer, D.W.%
, Collier, W.%
\BCBL {} Ruis, A.R.%
\end{APACrefauthors}%
\unskip\
\newblock
\APACrefYearMonthDay{2016}{}{}.
\newblock
{\BBOQ}\APACrefatitle {A tutorial on epistemic network analysis: Analyzing the structure of connections in cognitive, social, and interaction data} {A tutorial on epistemic network analysis: Analyzing the structure of connections in cognitive, social, and interaction data}.{\BBCQ}
\newblock
\APACjournalVolNumPages{Journal of Learning Analytics}{3}{3}{9--45,}
\newblock
\begin{APACrefDOI} \doi{10.18608/jla.2016.33.3} \end{APACrefDOI}
\newblock

\newblock

\PrintBackRefs{\CurrentBib}

\bibitem [\protect \citeauthoryear {%
Shea%
\ \BBA {} Chan%
}{%
Shea%
\ \BBA {} Chan%
}{%
{\protect \APACyear {2023}}%
}]{%
shea2023clinical}
\APACinsertmetastar {%
shea2023clinical}%
\begin{APACrefauthors}%
Shea, G.K.%
\BCBT {}\ \BBA {} Chan, P.C.%
\end{APACrefauthors}%
\unskip\
\newblock
\APACrefYearMonthDay{2023}{}{}.
\newblock
{\BBOQ}\APACrefatitle {Clinical reasoning in medical education: a primer for medical students} {Clinical reasoning in medical education: a primer for medical students}.{\BBCQ}
\newblock
\APACjournalVolNumPages{Teaching and Learning in Medicine}{36}{4}{547--555,}
\newblock
\begin{APACrefDOI} \doi{10.1080/10401334.2023.2230201} \end{APACrefDOI}
\newblock

\newblock

\PrintBackRefs{\CurrentBib}

\bibitem [\protect \citeauthoryear {%
Si%
}{%
Si%
}{%
{\protect \APACyear {2022}}%
}]{%
si2022strategies}
\APACinsertmetastar {%
si2022strategies}%
\begin{APACrefauthors}%
Si, J.%
\end{APACrefauthors}%
\unskip\
\newblock
\APACrefYearMonthDay{2022}{}{}.
\newblock
{\BBOQ}\APACrefatitle {Strategies for developing pre-clinical medical students' clinical reasoning based on illness script formation: a systematic review} {Strategies for developing pre-clinical medical students' clinical reasoning based on illness script formation: a systematic review}.{\BBCQ}
\newblock
\APACjournalVolNumPages{Korean Journal of Medical Education}{34}{1}{49--61,}
\newblock
\begin{APACrefDOI} \doi{10.3946/kjme.2022.219} \end{APACrefDOI}
\newblock

\newblock

\PrintBackRefs{\CurrentBib}

\bibitem [\protect \citeauthoryear {%
Sonnenberg%
\ \BBA {} Bannert%
}{%
Sonnenberg%
\ \BBA {} Bannert%
}{%
{\protect \APACyear {2015}}%
}]{%
sonnenberg2015discovering}
\APACinsertmetastar {%
sonnenberg2015discovering}%
\begin{APACrefauthors}%
Sonnenberg, C.%
\BCBT {}\ \BBA {} Bannert, M.%
\end{APACrefauthors}%
\unskip\
\newblock
\APACrefYearMonthDay{2015}{}{}.
\newblock
{\BBOQ}\APACrefatitle {Discovering the effects of metacognitive prompts on the sequential structure of SRL-processes using process mining techniques} {Discovering the effects of metacognitive prompts on the sequential structure of srl-processes using process mining techniques}.{\BBCQ}
\newblock
\APACjournalVolNumPages{Journal of Learning Analytics}{2}{1}{72--100,}
\newblock

\newblock

\PrintBackRefs{\CurrentBib}

\bibitem [\protect \citeauthoryear {%
Swiecki%
\ \protect \BOthers {.}}{%
Swiecki%
\ \protect \BOthers {.}}{%
{\protect \APACyear {2022}}%
}]{%
swiecki2022assessment}
\APACinsertmetastar {%
swiecki2022assessment}%
\begin{APACrefauthors}%
Swiecki, Z.%
, Khosravi, H.%
, Chen, G.%
, Martinez-Maldonado, R.%
, Lodge, J.M.%
, Milligan, S.%
\BDBL {}Ga{\v{s}}evi{\'c}, D.%
\end{APACrefauthors}%
\unskip\
\newblock
\APACrefYearMonthDay{2022}{}{}.
\newblock
{\BBOQ}\APACrefatitle {Assessment in the age of artificial intelligence} {Assessment in the age of artificial intelligence}.{\BBCQ}
\newblock
\APACjournalVolNumPages{Computers and Education: Artificial Intelligence}{3}{}{100075,}
\newblock

\newblock

\PrintBackRefs{\CurrentBib}

\bibitem [\protect \citeauthoryear {%
Tao%
, Song%
\BCBL {}\ \BBA {} Fu%
}{%
Tao%
\ \protect \BOthers {.}}{%
{\protect \APACyear {2025}}%
}]{%
tao2025exploring}
\APACinsertmetastar {%
tao2025exploring}%
\begin{APACrefauthors}%
Tao, L.%
, Song, Y.%
\BCBL {} Fu, J.%
\end{APACrefauthors}%
\unskip\
\newblock
\APACrefYearMonthDay{2025}{}{}.
\newblock
{\BBOQ}\APACrefatitle {Exploring students’ self-regulated learning behavioural patterns and perceptions in an English speaking task within a generative AI-supported immersive VR} {Exploring students’ self-regulated learning behavioural patterns and perceptions in an english speaking task within a generative ai-supported immersive vr}.{\BBCQ}
\newblock
\APACjournalVolNumPages{Computers \& Education}{}{}{105515,}
\newblock

\newblock

\PrintBackRefs{\CurrentBib}

\bibitem [\protect \citeauthoryear {%
Thampy%
, Willert%
\BCBL {}\ \BBA {} Ramani%
}{%
Thampy%
\ \protect \BOthers {.}}{%
{\protect \APACyear {2019}}%
}]{%
thampy2019assessing}
\APACinsertmetastar {%
thampy2019assessing}%
\begin{APACrefauthors}%
Thampy, H.%
, Willert, E.%
\BCBL {} Ramani, S.%
\end{APACrefauthors}%
\unskip\
\newblock
\APACrefYearMonthDay{2019}{}{}.
\newblock
{\BBOQ}\APACrefatitle {Assessing clinical reasoning: targeting the higher levels of the pyramid} {Assessing clinical reasoning: targeting the higher levels of the pyramid}.{\BBCQ}
\newblock
\APACjournalVolNumPages{Journal of general internal medicine}{34}{8}{1631--1636,}
\newblock

\newblock

\PrintBackRefs{\CurrentBib}

\bibitem [\protect \citeauthoryear {%
Veenman%
}{%
Veenman%
}{%
{\protect \APACyear {2007}}%
}]{%
veenman2007assessment}
\APACinsertmetastar {%
veenman2007assessment}%
\begin{APACrefauthors}%
Veenman, M.V.%
\end{APACrefauthors}%
\unskip\
\newblock
\APACrefYearMonthDay{2007}{}{}.
\newblock
{\BBOQ}\APACrefatitle {The assessment and instruction of self-regulation in computer-based environments: a discussion} {The assessment and instruction of self-regulation in computer-based environments: a discussion}.{\BBCQ}
\newblock
\APACjournalVolNumPages{Metacognition and Learning}{2}{2}{177--183,}
\newblock
\begin{APACrefDOI} \doi{10.1007/s11409-007-9017-6} \end{APACrefDOI}
\newblock

\newblock

\PrintBackRefs{\CurrentBib}

\bibitem [\protect \citeauthoryear {%
Verbert%
\ \protect \BOthers {.}}{%
Verbert%
\ \protect \BOthers {.}}{%
{\protect \APACyear {2014}}%
}]{%
verbert2014learning}
\APACinsertmetastar {%
verbert2014learning}%
\begin{APACrefauthors}%
Verbert, K.%
, Govaerts, S.%
, Duval, E.%
, Santos, J.L.%
, Van~Assche, F.%
, Parra, G.%
\BCBL {} Klerkx, J.%
\end{APACrefauthors}%
\unskip\
\newblock
\APACrefYearMonthDay{2014}{}{}.
\newblock
{\BBOQ}\APACrefatitle {Learning dashboards: an overview and future research opportunities} {Learning dashboards: an overview and future research opportunities}.{\BBCQ}
\newblock
\APACjournalVolNumPages{Personal and Ubiquitous Computing}{18}{6}{1499--1514,}
\newblock

\newblock

\PrintBackRefs{\CurrentBib}

\bibitem [\protect \citeauthoryear {%
Wagner-Menghin%
, de Bruin%
\BCBL {}\ \BBA {} van Merri{\"e}nboer%
}{%
Wagner-Menghin%
\ \protect \BOthers {.}}{%
{\protect \APACyear {2020}}%
}]{%
wagner2020communication}
\APACinsertmetastar {%
wagner2020communication}%
\begin{APACrefauthors}%
Wagner-Menghin, M.%
, de Bruin, A.B.%
\BCBL {} van Merri{\"e}nboer, J.J.%
\end{APACrefauthors}%
\unskip\
\newblock
\APACrefYearMonthDay{2020}{}{}.
\newblock
{\BBOQ}\APACrefatitle {Communication skills supervisors’ monitoring of history-taking performance: an observational study on how doctors and non-doctors use cues to prepare feedback} {Communication skills supervisors’ monitoring of history-taking performance: an observational study on how doctors and non-doctors use cues to prepare feedback}.{\BBCQ}
\newblock
\APACjournalVolNumPages{BMC medical education}{20}{1}{36,}
\newblock

\newblock

\PrintBackRefs{\CurrentBib}

\bibitem [\protect \citeauthoryear {%
Windish%
, Price%
, Clever%
, Magaziner%
\BCBL {}\ \BBA {} Thomas%
}{%
Windish%
\ \protect \BOthers {.}}{%
{\protect \APACyear {2005}}%
}]{%
windish2005teaching}
\APACinsertmetastar {%
windish2005teaching}%
\begin{APACrefauthors}%
Windish, D.M.%
, Price, E.G.%
, Clever, S.L.%
, Magaziner, J.L.%
\BCBL {} Thomas, P.A.%
\end{APACrefauthors}%
\unskip\
\newblock
\APACrefYearMonthDay{2005}{}{}.
\newblock
{\BBOQ}\APACrefatitle {Teaching medical students the important connection between communication and clinical reasoning} {Teaching medical students the important connection between communication and clinical reasoning}.{\BBCQ}
\newblock
\APACjournalVolNumPages{Journal of General Internal Medicine}{20}{12}{1108--1113,}
\newblock

\newblock

\PrintBackRefs{\CurrentBib}

\bibitem [\protect \citeauthoryear {%
Winne%
}{%
Winne%
}{%
{\protect \APACyear {2022}}%
}]{%
winne2022modeling}
\APACinsertmetastar {%
winne2022modeling}%
\begin{APACrefauthors}%
Winne, P.H.%
\end{APACrefauthors}%
\unskip\
\newblock
\APACrefYearMonthDay{2022}{}{}.
\newblock
{\BBOQ}\APACrefatitle {Modeling self-regulated learning as learners doing learning science: How trace data and learning analytics help develop skills for self-regulated learning} {Modeling self-regulated learning as learners doing learning science: How trace data and learning analytics help develop skills for self-regulated learning}.{\BBCQ}
\newblock
\APACjournalVolNumPages{Metacognition and Learning}{17}{3}{773--791,}
\newblock

\newblock

\PrintBackRefs{\CurrentBib}

\bibitem [\protect \citeauthoryear {%
Woodham%
\ \protect \BOthers {.}}{%
Woodham%
\ \protect \BOthers {.}}{%
{\protect \APACyear {2019}}%
}]{%
woodham2019virtual}
\APACinsertmetastar {%
woodham2019virtual}%
\begin{APACrefauthors}%
Woodham, L.A.%
, Round, J.%
, Stenfors, T.%
, Bujacz, A.%
, Karlgren, K.%
, Jivram, T.%
\BDBL {}Poulton, T.%
\end{APACrefauthors}%
\unskip\
\newblock
\APACrefYearMonthDay{2019}{}{}.
\newblock
{\BBOQ}\APACrefatitle {Virtual patients designed for training against medical error: Exploring the impact of decision-making on learner motivation} {Virtual patients designed for training against medical error: Exploring the impact of decision-making on learner motivation}.{\BBCQ}
\newblock
\APACjournalVolNumPages{PloS one}{14}{4}{e0215597,}
\newblock

\newblock

\PrintBackRefs{\CurrentBib}

\bibitem [\protect \citeauthoryear {%
Xu%
, Ang%
, Soh%
\BCBL {}\ \BBA {} Ponnamperuma%
}{%
Xu%
\ \protect \BOthers {.}}{%
{\protect \APACyear {2021}}%
}]{%
xu2021methods}
\APACinsertmetastar {%
xu2021methods}%
\begin{APACrefauthors}%
Xu, H.%
, Ang, B.W.G.%
, Soh, J.Y.%
\BCBL {} Ponnamperuma, G.G.%
\end{APACrefauthors}%
\unskip\
\newblock
\APACrefYearMonthDay{2021}{}{}.
\newblock
{\BBOQ}\APACrefatitle {Methods to improve diagnostic reasoning in undergraduate medical education in the clinical setting: a systematic review} {Methods to improve diagnostic reasoning in undergraduate medical education in the clinical setting: a systematic review}.{\BBCQ}
\newblock
\APACjournalVolNumPages{Journal of General Internal Medicine}{36}{9}{2745--2754,}
\newblock
\begin{APACrefDOI} \doi{10.1007/s11606-021-06916-0} \end{APACrefDOI}
\newblock

\newblock

\PrintBackRefs{\CurrentBib}

\bibitem [\protect \citeauthoryear {%
Yi%
\ \BBA {} Kim%
}{%
Yi%
\ \BBA {} Kim%
}{%
{\protect \APACyear {2025}}%
}]{%
yi2025feasibility}
\APACinsertmetastar {%
yi2025feasibility}%
\begin{APACrefauthors}%
Yi, Y.%
\BCBT {}\ \BBA {} Kim, K\BHBI J.%
\end{APACrefauthors}%
\unskip\
\newblock
\APACrefYearMonthDay{2025}{}{}.
\newblock
{\BBOQ}\APACrefatitle {The feasibility of using generative artificial intelligence for history taking in virtual patients} {The feasibility of using generative artificial intelligence for history taking in virtual patients}.{\BBCQ}
\newblock
\APACjournalVolNumPages{BMC Research Notes}{18}{1}{80,}
\newblock

\newblock

\PrintBackRefs{\CurrentBib}

\end{thebibliography}
